\documentstyle[jair,twoside,11pt,theapa]{article}
\input epsf

\jairheading{26}{2006}{289-322}{10/04}{07/06}
\ShortHeadings{Breaking Instance-Independent Symmetries 
 in Exact Graph Coloring}
{Ramani, Aloul, Markov, \& Sakallah}
\firstpageno{289}

\title{Breaking Instance-Independent Symmetries \\ in Exact Graph Coloring}

\author{\name Arathi Ramani \email ramania@umich.edu \\
       \name Igor L. Markov \email imarkov@eecs.umich.edu \\
       \name Karem A. Sakallah \email karem@eecs.umich.edu \\
       \addr Department of Electrical Engineering and Computer Science\\
       University of Michigan, Ann Arbor, USA
       \AND
       \name Fadi A. Aloul \email faloul@umich.edu \\
       \addr Department of Computer Engineering\\ 
       American University in Sharjah, UAE}

\begin{document}

\maketitle

\begin{abstract}
   Code optimization and high level synthesis can be posed as
   constraint satisfaction and optimization problems,
   such as graph coloring used in register allocation.
   Graph coloring is also used to model more traditional
   CSPs relevant to AI, such as planning, time-tabling and scheduling.
   Provably optimal solutions may be desirable for commercial
   and defense applications. Additionally, for applications such as
   register allocation and code optimization, naturally-occurring instances 
   of graph coloring are often small and can be solved optimally. A recent 
   wave of improvements in algorithms for Boolean satisfiability (SAT) and
   0-1 Integer Linear Programming (ILP) suggests generic problem-reduction
   methods, rather than problem-specific heuristics, because (1) heuristics
   may be upset by new constraints, (2) heuristics tend to ignore structure,
   and (3) many relevant problems are provably inapproximable. 
  
   Problem reductions often lead to highly symmetric SAT instances,
   and symmetries are known to slow down SAT solvers. In this work,
   we compare several avenues for symmetry breaking, in particular when
   certain kinds of symmetry are present in all generated instances. 
   Our focus on reducing CSPs to SAT allows us to leverage
   recent dramatic improvement in SAT solvers and automatically
   benefit from future progress. We can use a variety of black-box SAT solvers 
   without modifying their source code because our symmetry-breaking techniques
   are static, i.e., we detect symmetries and add symmetry breaking predicates
   (SBPs) during pre-processing.

   An important result of our work is that among the types 
   of instance-independent SBPs we studied and their combinations,
   the simplest and least complete constructions are the most effective.
   Our experiments also clearly indicate that instance-independent
   symmetries should mostly be processed together with instance-specific
   symmetries rather than at the specification level, contrary 
   to what has been suggested in the literature. 
\end{abstract}
   
\section{Introduction}
\label{sec:intro} 

   Detecting and using problem structure, such as symmetries,
   can often be very useful in accelerating the search for solutions of  
   constraint satisfaction problems (CSPs).
   This is particularly true for algorithms which perform exhaustive
   searches and benefit from pruning the search tree.
   This work conducts a theoretical and empirical study of the impact
   of breaking structural symmetries in 0-1 ILP reductions of
   the exact graph coloring problem  which has applications 
   in a number of fields.
   For example, in compiler design, many techniques for code optimization 
   and high-level synthesis 
   operate with relatively few objects at a time.  
   Graph coloring used for register allocation during program compilation
   \cite{Chaitin81} is limited by small numbers of registers in 
   embedded processors as well as
   by the number of local variables and virtual registers. 
   Graph coloring is  also relevant to
   AI applications such as
   planning, scheduling, and map coloring. Recent work on graph coloring
   in AI has included algorithms based on neural networks \cite{Jagota96},
   evolutionary algorithms \cite{Gal99}, scatter search  \cite{Hamiez01}
   and several other approaches discussed in Section \ref{sec:bkg}. 
   While many of these search procedures
   are heuristic, our work focuses on exact graph coloring, 
   which is closely related to  several useful combinatorial problems
   such as {\em maximal independent set} and {\em vertex cover}.
   We seek provably optimal solutions because they may be desirable 
   in commercial
   and defense applications for competitive reasons,
   and can often be found.
   Our work focuses on solving exact graph coloring by reduction to 0-1 ILP.
   While the idea of solving ${\mathcal NP}-complete$ problems
   by reduction is well-known, it is rarely used in practice 
   because algorithms developed for ``standard'' problems, such as SAT, 
   may not be competitive with domain-specific techniques
   that are aware of problem structure. However, many applications 
   imply problem-specific constraints and non-trivial 
   objective functions. These extensions may upset heuristics 
   for standard problems. Heuristics, particularly those based on local search,
   often fail to use structure in problem instances \cite{Prestwich02}
   and are inefficient when used with problem reductions.
   In contrast, exact solvers based on branch-and-bound and back-tracking 
   tend to adapt to new constraints and can be applied through 
   problem reduction. There is a growing literature on handling structure
   in optimal solvers \cite{Aloul02a,Crawford96,Huang03}, and our work 
   falls into this category as well. 

     The NP-spec project \cite{Cadoli99} offers a framework for 
     formulating a wide range of combinatorial problems 
 and automatically reducing 
   their instances to instances of Boolean satisfiability.
 This approach is attractive because it circumvents 
   problem-specific solvers and leverages recent breakthroughs
   in Boolean satisfiability \cite{Chaff}. However, this approach 
   remains unexplored in practice, possibly because the 
  efficiency 
   of problem-solving may be reduced when domain-specific structure is lost
   during problem reductions. This drawback is addressed
   by recent work on the detection of structure, particularly symmetry,
   in SAT and 0-1 ILP instances in order 
   to accelerate exact solvers \cite{Crawford96,Aloul02a,Aloul04}. In these
   papers, symmetries in a SAT/0-1 ILP instance are detected by reduction to
   {\em graph automorphism}, i.e. the formula is represented by a graph
   and the automorphism problem for that graph is solved using 
   graph automorphism
   software packages \cite{Nauty,Saucy}. Until recently, this type of
symmetry detection was frequently inefficient because 
 solving the  automorphism problem for large graphs can be time-consuming. 
 However, more recent automorphism software \cite{Saucy} 
has removed this bottleneck to a large extent.
   Moreover, adding simple symmetry breaking predicates 
   as new constraints significantly speeds up exact SAT solvers \cite{Aloul02a}.
      This work can be viewed as a case study of symmetry breaking 
   in problem reductions, as we focus on {\em graph coloring} and its variants
   that can be reduced to Boolean satisfiability and 0-1 ILP.
   Our main goals are to (i) accelerate optimal solving of graph coloring
   instances, and (ii) compare different strategies for breaking 
   instance-independent symmetries. 
   There are two distinct sources of symmetries in graph-coloring instances:
   (i) colors can be arbitrarily permuted (instance-independent symmetries),
   and (ii) some graphs may be invariant under certain permutations of vertices
   (instance-dependent symmetries). Previous work \cite{Crawford96,Aloul02a,Aloul04}
   deals only with instance-dependent symmetries in SAT and 0-1 ILP instances.
   Symmetries are first detected by reduction to graph automorphism and then
   broken by adding symmetry breaking predicates (SBPs) to the formulation.
  The advantage of such a strategy  is that every instance-independent symmetry is also
     instance-dependent, whereas the reverse does not hold. Symmetries that exist due to problem
     formulation appear in every instance of the problem, in addition to symmetries that exist due
     to specific parameter values for an instance.  
   Given that there may be many instance-specific symmetries,
   one may process all symmetries at once using publicly available 
   symmetry processing packages such as {\sc Shatter} \cite{Aloul02a,Aloul03a}. 
   Alternatively, one may add symmetry breaking
   predicates for instance-independent symmetries early, hoping to 
   speed-up the processing of remaining symmetries. This type
   of symmetry breaking has not been discussed in earlier work
    \cite{Aloul02a,Aloul03a}, and in this paper we study its
   utility for the graph coloring problem.

 Our work deals with symmetries of problem and instance descriptions;
 we distinguish (i) symmetries of generic problem specifications from 
 (ii) symmetries of problem-instance data. The former symmetries
 translate to the latter but not the other way around --- an example
 from graph coloring is given by color permutations versus automorphisms
 of specific graphs. While both types of symmetries
 can be detected by solving the graph automorphism problem, symmetries 
 in specifications can often be captured manually, whereas capturing 
 symmetries in problem instances may require large-scale computation
 and non-trivial software. Indeed, when specification-level symmetries
 are instantiated, the size of their support (the number of objects moved)
 typically increases dramatically. For example, color permutations in
 graph coloring should be simultaneously applied to every vertex 
 of a graph in question.
 Detecting symmetries with larger support seems like a waste of computational
 effort. To this end, recent work on breaking symmetries in specifications \cite{Cadoli03} 
 prefers instance-independent 
techniques and breaks symmetries only at the specification level.
This approach is particularly relevant with constraint solvers and
languages that process problem specifications prior to seeing actual
problem instances and can amortize the symmetry-detection effort. Also,
in a more general setting, using instance-independent symmetry breaking 
does not rule out applying redundant (or complementary) instance-specific 
techniques at a later stage.

 Until recently automatic symmetry detection had been a serious
bottleneck in handling symmetries. For example, if graph automorphism
is solved using the program {\sc Nauty} \cite{Nauty}, detecting symmetries 
often can take longer than constraint solving without symmetry breaking. 
This was observed for microprocessor verification SAT instances 
 by Aloul et. al. in 2002 \cite{Aloul02a}. Therefore, detecting symmetries early and representing
them in a more structured way appears attractive, especially given that 
this may potentially increase the efficiency of symmetry-breaking. 
However, the symmetry-detection bottleneck has recently been eliminated
in many applications with the software tool {\sc Saucy} \cite{Saucy} that
often finds symmetries of practical graphs many times faster than {\sc Nauty}. 
This development undermines, to some extent, the potential benefits 
of symmetry processing at the specification level and puts 
the spotlight on symmetry-breaking. To that end, SBPs 
  added in different circumstances
may have different efficiency, and while it is unclear {\em a priori}
which approach is more successful, the differences in performance may be
significant. Since SBPs appear to the solver as additional constraints, 
they may either speed up or frustrate the solver
(the latter effect is clearly visible in our experiments with CPLEX).
Outcomes of practical experiments are also affected by recent dramatic
improvements in the efficiency of symmetry-breaking predicates
\cite{Aloul03a,Aloul04}. While it seems difficult to justify 
any particular expectation for empirical performance, we are fortunate
to observe clear trends in experimental data presented in 
Section \ref{sec:results} and summarize them with simple rules.


    While we focus on graph coloring instances, our techniques
    are immediately applicable to related CSP problems, e.g.,
    those produced by adding new types of constraints that can be
    easily expressed in SAT or 0-1 ILP when graph coloring is converted
    to those generic problems. We also expect that our conclusions
    about symmetry-breaking carry over to other CSPs that can be
    economically reduced to SAT and 0-1 ILP, e.g., maximum independent set,
    minimum dominating set, etc. Another advantage of our approach is
    being able to use a variety of existing and future SAT and 0-1 ILP
    solvers without modifying their source code. Unfortunately, this
    precludes the use of dynamic symmetry-breaking that would
    require modifying the source code and may adversely affect
    performance by disturbing the fragile balance between
    the amount of reasoning and searching performed by modern SAT solvers. 
    Specifically, heuristics for variable ordering and decision selection 
    may be affected, as well as the recording of learned conflict clauses
    (nogoods).

 The main contributions of this work are listed below.

\begin{itemize}
\item Using the symmetry breaking flow for pseudo-Boolean (PB) formulas
described by Aloul et. al in 2004 \cite{Aloul04}, we detect and break symmetries
in DIMACS graph coloring benchmarks expressed as instances of 0-1 
ILP. We show that instance-dependent symmetry breaking
enables many medium-sized instances to be optimally solved 
in reasonable time on commodity PCs 
\item We propose  instance-independent techniques
for breaking symmetries during problem formulation, 
assess their relative strength and completeness, and evaluate
them empirically using well-known academic and commercial tools

\item We show empirically that instance-dependent techniques
are, in general, more effective than instance-independent symmetry
breaking for the benchmarks in question. In fact, only
the simplest and least complex instance-independent SBPs
are competitive 
\end{itemize}

 The remaining part of the paper is organized as follows.
Section \ref{sec:bkg} covers background 
on graph coloring, SAT and 0-1 ILP, as well as
previous work on symmetry breaking.
Instance-independent symmetry breaking predicates
are discussed in Section \ref{sec:sbconst}.  Section \ref{sec:results} 
presents our empirical results and Section \ref{sec:conclude} concludes
the paper. The Appendix gives detailed results for the {\tt queens} 
family of instances.

\section{Background and Previous Work}
\label{sec:bkg}
This section discusses problem definitions and applications of
some existing algorithms for exact graph coloring.
We also discuss previous work on symmetry breaking
for SAT and 0-1 ILP in some detail.

\subsection{Graph Coloring}

Given an undirected graph $G(V,E)$, a {\bf vertex coloring} of the graph 
 is an assignment of a label (color) to each node such that
the labels on adjacent nodes are different. A minimum coloring 
uses the smallest possible number of colors, known as the 
{\em chromatic number} of a graph.
The {\em decision version} of graph coloring ($K-$coloring)
asks whether vertices in a graph can be colored using $\leq K$ colors
for a given $K$.

A {\em clique} of an undirected graph $G(V,E)$ is a set of 
mutually adjacent vertices in the graph.
The {\bf maximum clique} problem consists of seeking a clique of maximal size, 
i.e., a clique with at least as many vertices as any other clique in
the graph. The maximum clique and graph coloring problems
are closely related. Specifically, the max-clique size is a lower bound 
on the chromatic number of the graph. 
 Over the years, a number of different algorithms for
solving graph coloring have been developed, because of
its fundamental importance in computer science. 
These algorithms fall into three broad categories: 
polynomial-time approximation schemes, optimal algorithms, and heuristics.
We briefly discuss work in each of these categories below.
There are a number of online resources on graph coloring \cite{Trick,Culberson} that offer more detailed bibliographies.

As far as approximation schemes are concerned, the
most common technique used is {\em successive augmentation}.
In this approach a partial coloring is found on a small number 
of vertices and this is extended vertex by vertex until the entire graph is colored.
Examples include the algorithms by 
Leighton \cite{Leighton79} for large scheduling problems,
and by Welsh and Powell \cite{Welsh67} for time-tabling.
More recent work has attempted to tighten the worst-case
bounds on the chromatic number of the graph. The algorithm providing the currently best worst-case ratio 
(number of colors used divided by optimal number) is due to
Haldorsson \cite{Hald90}, and guarantees a ratio
of no more than $O\left( \frac{n{(\log \log n)}^2}{{(\log n)}^3} \right)$,
where $n$ is the number of vertices.
General heuristic methods that have been tried include
simulated annealing \cite{Chams87,Johnson91} and tabu search
\cite{Hertz87}. A well-known heuristic that is still
widely used is the DSATUR algorithm by Brelaz \cite{Brelaz79} 
which colors vertices according to their {\em saturation degree}.
The saturation degree of a vertex is the number of
different colors to which it is adjacent. The DSATUR
heuristic repeatedly picks a vertex with maximal
saturation degree and colors it with the lowest-numbered
color possible. This heuristic is optimal
for bipartite graphs. Algorithms for finding optimal colorings 
are frequently based on implicit enumeration, and are discussed in
more detail later in this section. Both the graph
coloring and max-clique problems are ${\mathcal NP}$-complete
\cite{GareyNP} and even finding near-optimal solutions
with good approximation guarantees is ${\mathcal NP}$-hard \cite{Feige91}. 
The inapproximability of graph coloring suggests that it may be
more difficult to solve heuristically than, say, the
Traveling Salesman Problem for which Polynomial-Time Approximation
Schemes (PTAS) are known for Euclidean and Manhattan graphs. 
For this and a number of other reasons, 
we study optimal graph coloring and many application-derived instances
that are solvable in reasonable time.  Several applications are outlined next.

{\sc  Time-Tabling and Scheduling}
problems involve placing pairwise restrictions
on jobs that cannot be performed simultaneously.
For example, two classes taught by the same
faculty member cannot be scheduled in the same time slot.
The problem has been studied in previous work by Leighton \cite{Leighton79}
and De Werra \cite{Dewerra85}. 
More generally, graph coloring is an important problem in 
Artificial Intelligence
because of its close relationship to planning and scheduling.
Several traditional AI techniques have been applied to this problem, including 
 parallel algorithms using neural networks \cite{Jagota96}. 
Genetic and hybrid evolutionary algorithms have also been developed,  
notably by Galinier et. al. in 1999 \cite{Gal99}, in addition to more traditional optimization
methodology, such as scatter search \cite{Hamiez01}. There have also been studies of
benchmarking models for graph coloring, such as the recent work by Walsh 
\cite{Walsh01}, which shows that graphs with high vertex degrees are more likely to
occur in real-world applications.

{\sc Register Allocation} is a very active application
of graph coloring.
This problem seeks to assign variables
to a limited number of hardware registers during program 
execution. Accessing variables in registers is much faster
than fetching them from memory. However, the number of registers
is limited and is typically much smaller than the number
of variables. Therefore, multiple variables must be
assigned to the same register. There are restrictions
on these assignments.  Two variables conflict with each other if 
they are {\em live} at the same time, i.e.
one is used both before and after the other within a 
short period of time (for instance, within a subroutine). 
The goal is to assign variables that do not conflict so as 
to minimize the use of non-register memory.
To formalize this, one creates a graph where 
nodes represent variables and edges represent conflicts 
between variables. A coloring maps to a conflict-free 
assignment, and if the number of registers exceeds the chromatic number,
a conflict-free register assignment exists \cite{Chaitin81}.

{\sc Printed Circuit Board Testing} \cite{GareyNP} involves the problem of 
testing printed circuit boards (PCBs)
for unintended short circuits (caused by stray lines of solder). 
This gives rise to a graph coloring problem in which the vertices 
correspond to the nets on board and there is an edge between two 
vertices if there is a potential for a short circuit between the 
corresponding nets. Coloring the graph corresponds to partitioning 
the nets into ``supernets,'' where the nets in each supernet 
can be simultaneously tested for shorts against all other nets, 
thereby speeding up the testing process.

{\sc Radio frequency assignment}  for  broadcast services in geographic regions
   (including commercial radio stations, taxi dispatch, police and emergency services).  
   The list of all possible frequencies is fixed by government agencies, but adjacent 
   geographic regions cannot use
   overlapping frequencies. To reduce frequency assignment to graph coloring,
    each geographic region needing K frequencies is represented with a $K-$clique, 
   and all $N \times K$ possible bipartite edges are introduced between two geographically
   adjacent regions needing $N$ and $K$ frequencies respectively.

Other applications of graph coloring in circuit design
and layout include circuit clustering, scheduling for
signal flow graphs, and many others. Benchmarks from 
these applications are not publicly available, and
therefore do not appear in this paper. However, all the
symmetry breaking   techniques described here 
extend to instances from any application. The benchmarks
we use here do include register allocation,
$n-$queens, and several other applications discussed in more 
detail in Section \ref{sec:results}. Empirically,
we observe that many of the instances in this paper can be
optimally solved in reasonable time, especially when symmetry breaking
is employed. Since this work deals with finding optimal solutions
for graph coloring, we discuss previous work on finding
exact algorithms for this problem in some detail.

The literature on exact graph coloring 
includes generic algorithms  \cite{Kubale85}
and specialized algorithms for a
particular application, such as Chaitin's
register allocation algorithm \cite{Chaitin81}.
At the moment, there does not appear to be a comprehensive
survey of techniques for this problem. However, 
online surveys \cite{Trick,Culberson} contain 
reasonably large bibliographies and even downloadable
source code for coloring algorithms in some cases.
Published algorithms for finding optimal graph colorings are mainly based
on implicit enumeration. The algorithm proposed by 
Brown \cite{Brown72} enumerates solutions for a given
instance of graph coloring and checks each solution
for correctness and optimality. The algorithm introduces
a special {\em tree construction} to avoid redundancy in
enumerating solutions. The work by Brelaz \cite{Brelaz79} improves upon
this algorithm by creating an initial coloring
based on some clique in the graph and then considering assignments induced by
this coloring. The work by Kubale and Kusz \cite{Kubale83} discusses the empirical
performance of implicit enumeration algorithms, and later work by Kubale and Jackowski 
 \cite{Kubale85} augments traditional 
implicit enumeration techniques with more sophisticated
backtracking methods.

Our work deals with solving graph coloring by {\em reduction} 
to another problem, in this case 0-1 ILP. This type of
reduction has been discussed in the past, notably in the recent work by Mehrotra and Trick 
\cite{Meh96}, which proposes an optimal coloring algorithm 
which expresses graph coloring using ILP-like constraints.
It relies on an auxiliary independent set formulation,
where each independent set in a graph is represented by a variable.
There can be prohibitively many variables but in practical cases 
this number may be reduced by {\em column generation}, a method 
that first tries to solve a linear relaxation using
a subset of variables and then adds more where needed.
This approach inherently breaks problem symmetries,
and thus rules out the use of SBPs 
as a way to speed up the search process. Our ILP construction
differs considerably from the one described above, since it does
not rely on an independent set formulation, but assigns
colors to individual vertices by using indicator variables.
The construction is described in more detail later in this section.
Solving graph coloring by reduction allows exact solutions to be found
by using SAT/0-1 ILP solvers as black boxes. Earlier work by Coudert  
 \cite{Coudert97} demonstrated that finding exact solutions for application-derived 
graph coloring benchmarks often takes no longer than heuristic approaches,
and that heuristic solutions may differ from the optimal value by as much as 
100\%. Coudert \cite{Coudert97} proposes an algorithm that finds exact 
graph coloring solutions by solving the max-clique problem. The algorithm uses
a technique called ``$q-$color pruning'', which assigns colors to vertices 
and systematically removes vertices that can be colored by $q$ colors, 
where $q$ is greater than a specified limit.

\subsection{Breaking Symmetries in CSPs}
\label{subsec:symcsps}

Several earlier works have addressed the importance of symmetry breaking
in the search for solutions of CSPs. It has been shown \cite{Krish85} that 
symmetry facilitates short proofs of propositions such as the pigeonhole
principle, whereas pure-resolution proofs are necessarily exponential in size.
Finding such proofs is, of course, a very difficult problem, but the performance
of many CSP techniques can be lower-bounded by the best-case proof size.
A typical approach to use symmetries is to prevent a CSP solver from
considering redundant symmetric solutions. This is called symmetry-breaking
and can be accomplished by adding constraints, often called
symmetry-breaking predicates (SBPs).  {\em Static} symmetry-breaking, 
such as the instance-independent constructions proposed in this work and the instance-dependent 
predicates from the literature \cite{Aloul02a,Crawford96}, detects symmetries and adds SBPs
during pre-processing and not when branching toward possible solutions.
The Symmetry Breaking by Dominance Detection (SBDD) procedure described by Fahle
 in 2001 \cite{Fahle01} detects symmetric choice points during search. 
Each choice point generated by the search algorithm is checked against
previously expanded search nodes. If the same or an equivalent choice point
has been previously expanded, the choice point is not visited again.
The global cut algorithm proposed by Focacci and Milano \cite{Foc01} records all nogoods found during search
whose symmetric images should be pruned. This set of nogoods, called the
``global cut seed'' is used to generate global cut constraints that prune
symmetric images for the entire search tree, while ensuring that correctness of
the original constraints is not violated.  Later work \cite{Puget02} has proposed 
 improved methods for nogood recording. These works
do not offer a systematic strategy for symmetry detection - they either require
symmetries to be known or declared in advance, or record information during
search that enables symmetry detection. Our work outlines and implements
a complete strategy to detect and break symmetries automatically during 
pre-processing, so that a black-box solver can be used during search. 
This context is broader than those that justify the development of specialized
solvers.  On the other hand, our techniques do not conflict with dynamic
symmetry-breaking and some of our results can potentially be reused in that
context.

A promising new partially-dynamic approach to symmetry-breaking, called Group
Equivalence (GE) trees is proposed by Roney et. al. \cite{Roney04}. This work aims to reduce the
per-node overhead associated with dynamic approaches.  A GE tree is constructed
from a CSP with a symmetry group G such that the nodes of the tree represent
equivalence classes of partial assignments under the group. This approach is
illustrated  by tracking {\em value} symmetries, i.e.,
simultaneous permutations of values in CSP variables.  The work also shows that GE trees empirically outperform
several well-known symmetry-breaking methodologies, such as 
SBDDs.  In comparison, our work compares different ways to handle arbitrary
compositions of variable and value symmetries (in graph coloring,
value symmetries are seen at the specification level, whereas variable
symmetries can only be seen in problem instances).
To this end, our static techniques appear compatible rather than competing
with the use of GE trees. There have also been many symmetry breaking 
approaches with particular relevance to graph coloring. Recent work by Gent \cite{Gent01}
 proposes constraints that break symmetry between ``indistinguishable
values'', but does not evaluate them empirically. Like the lowest-index
ordering (LI) constraints proposed by us in Section \ref{sec:sbconst}, 
these constraints also use the pre-existing sequential numbering of vertices
in an instance of graph coloring to enforce distinctions between symmetric 
vertices. The construction appears complex compared to alternative SBPs and
not as effective in our experiments as simpler constructions. 
Another related work is \cite{Henten03}, which proposes a
constant-time, constant-space algorithm for detecting and breaking value
symmetries in a class of 
CSPs that includes graph coloring. More recently, Benhamou \cite{Benhamou04} discusses symmetry breaking for CSPs modeled
using not-equals constraints (NECSP), and uses graph coloring as an illustrative example. The paper defines
a sufficient condition for symmetry such that certain symmetries can be
detected in linear time. The removal of these symmetries leads to considerable
gains in backtracking search algorithms for NECSPs. In general, our
empirical results, reported in Section \ref{sec:results}, appear
competitive with those for state-of-the-art dynamic approaches. However,
designing the world's best graph-colorer is not the goal of our research.
Instead, we focus on more efficient problem reductions to SAT and 0-1 ILP
by improving symmetry-breaking. To ensure a broad applicability of our 
results, we treat SAT solvers as black boxes, and perform a comprehensive
comparison of static SBPs and report empirical trends. While a more 
comprehensive comparison against existing graph coloring literature
would be of great value, making it rigorous, conclusive and revealing 
requires that the best static and the best dynamic symmetry-breaking 
techniques are known. To this end, we speculate that a more likely winner
would be a hybrid. Additional major issues to be resolved include
the tuning of solvers to specific benchmarks (noted in the work by Kirovski and Potkonjak \cite{KirovskiP98},
differences in experimental setup, different software and hardware platforms,
etc. Given that such a comparison is not completely in the scope of our work,
it is better delegated to a dedicated publication. However, to demonstrate that our
techniques are competitive with related work, we provide a comparison with the best results 
 from recent literature \cite{Benhamou04,Coudert97} in Section \ref{subsec:relwork}.
%

\subsection{SAT and 0-1 ILP}
One can solve the decision version of graph coloring
 by reducing it to Boolean satisfiability, and the optimization
 version by reduction to 0-1 ILP.
 The Boolean satisfiability (SAT) problem 
involves finding an assignment to a set of {\em 0-1 variables}
that satisfies a set of constraints, called {\em clauses}, 
expressed in conjunctive normal 
form (CNF). A CNF formula on  $n$ binary variables, $x_1, \ldots ,x_n$
consists of a conjunction 
of  clauses, $\omega_1, \ldots ,
\omega_m$.  A clause consists of a disjunction
of {\em literals}. 
A literal $l$  is an occurrence of a Boolean variable or its complement.
The 0-1 ILP problem is closely related to SAT, and allows the use of
{\em pseudo-Boolean} (PB) constraints,  which are linear 
inequalities with integer coefficients that can be
expressed in the normalized form  \cite{Aloul02b} of:
$a_1x_1 \; + \; a_2x_2 \; + \; \ldots \; a_nx_n \; \leq \; b$
where $a_i, b \in Z^+$ and $x_i$ are literals of Boolean 
variables.\footnote{Using the relations $(Ax \geq b) \Leftrightarrow (-Ax \leq -b)$  and $\overline{x_i} = (1 - x_i)$, any arbitrary PB constraint 
can be expressed in normalized form with only positive coefficients.} 
In some cases a single PB constraint
can replace an exponential number of CNF clauses \cite{Aloul02b}.
In general, the efficiency of CNF reductions is encoding-dependent.
Earlier work by Warners \cite{Warners98} shows that a linear-overhead conversion exists
 from linear inequalities with integer coefficients and 0-1 variables
to CNF. However, CNF encodings which do not use this conversion may
be less efficient. When converting CNF to PB,  
a single CNF constraint can always be expressed as
a single 0-1 ILP constraint (by replacing disjunctions
between literals in the constraint with `+' and
setting the right-hand-side value as $\geq 1$). However,
this may not always be suitable since certain operations,
such as disjunction, implication and inequality are more
intuitively expressed as CNF, and can be efficiently processed
by SAT solvers such as Chaff \cite{Chaff}. A conversion to
0-1 ILP is more desirable for arithmetic operations, or
``counting constraints'', whose CNF equivalent requires
polynomially many clauses (and exponentially many for some conversions). To maximize the advantages of
both CNF and PB formats, most recent
0-1 ILP solvers such as PBS \cite{Aloul02b} and Galena \cite{Chai03}
allow a formula to possess  CNF {\em and} PB components.
Additionally, 0-1 ILP solvers also provide for the
solution of   { \em optimization
problems}. Subject to given constraints, one may request the minimization
(or maximization) of an objective function which must be a linear combination of the
problem variables.

Exact SAT solvers \cite{Berkmin,Chaff,Grasp}
are typically based on the original Davis-Logemann-Loveland (DLL) 
backtrack search algorithm 
\cite{Davis62}. Recently, several powerful methods have 
been proposed to expedite the backtrack search algorithm, such as 
conflict diagnosis \cite{Grasp} and watched literal Boolean 
constraint propagation (BCP) \cite{Chaff}. With these improvements,
modern SAT solvers  \cite{Chaff,Berkmin} are capable of solving instances with
several million variables and clauses in reasonable time.
This increase in scalability and scope has enabled
a number of SAT-based applications in various domains, including
circuit layout \cite{Aloul02a}, microprocessor verification,
symbolic model checking, and many others. More recent work has focused
on extending advances in SAT to 0-1 ILP \cite{Aloul02b,Chai03}. 
In this work, we focus on solving instances of exact graph coloring
by reduction to 0-1 ILP and the use of SBPs. Our choice of 0-1 ILP is motivated by the following reasons.

Firstly, 0-1 ILP permits the use
 of a more general input format than CNF, allowing greater efficiency in problem encoding,
but at the same time is similar enough to SAT to allow improved 
methods for SAT-solving to be used without paying a penalty for
generality. The specialized 0-1 ILP solvers PBS \cite{Aloul02b}
and Galena \cite{Chai03} both propose sophisticated new techniques
for 0-1 ILP that are based on recent decision
heuristics \cite{Chaff}, conflict diagnosis and backtracking
techniques \cite{Grasp} for SAT solvers. As a result, they
empirically perform better than both the generic ILP solver
CPLEX \cite{Cplex} and the leading-edge SAT solver zChaff on
several DIMACS SAT benchmarks and application-derived instances
such as FPGA routing instances from circuit layout.
Also, since 0-1 ILP is an optimization problem, unlike
SAT which is a decision problem, 0-1 ILP solvers possess 
the ability to maximize/minimize an objective function. They can, therefore,
be directly applied to the optimization version of exact
graph coloring, unlike pure CNF-SAT solvers that can only
be used on the $k-$coloring decision variant. It is possible
to solve the optimization version by repeatedly solving instances of the $k-$coloring
using a SAT solver, with the value of $k$ being updated after each call. However,
0-1  ILP solvers do not require this extra step, and moreover
tend to provide better performance than repeated calls to
a SAT solver on many Boolean optimization problems \cite{Aloul02b}.

It is possible to use a generic ILP solver, such as the
commercial solver CPLEX \cite{Cplex} instead of a specialized 0-1 ILP
solver without any changes in problem formulation. However, Aloul et al. \cite{Aloul02b} 
 show that this generalization is not always desirable, particularly
in the case of Boolean optimization problems such as Max-SAT.
0-1 ILP is also especially useful for evaluating the effectiveness
of symmetry breaking for graph coloring, the primary purpose of this work.
Detecting and breaking symmetries in SAT formulas has been
shown to speed up the problem-solving process \cite{Crawford96,Aloul02a}.
Recently, symmetry breaking techniques for SAT have been extended to
0-1 ILP \cite{Aloul04}, and have been shown to produce search speedups
in this domain as well. However, a similar extension for non-binary
variables for generic ILP does not presently exist. There is evidence \cite{Aloul02b} that
the advantages of symmetry breaking may depend on the
actual algorithm used in the search. Specifically, results in the cited work 
suggest that the generic ILP solver CPLEX is actually {\em slowed down}
by the addition of SBPs. Since CPLEX is a commercial tool and the algorithms
used by it are not publicly known, it is difficult to pinpoint 
a reason for this disparity. However, our empirical results in 
Section \ref{sec:results} do bear out these observations. 
The remainder of this section discusses the reduction of graph coloring
to 0-1 ILP and explains previous work in symmetry breaking in some detail.

\subsection {Detecting and Breaking Symmetries in 0-1 ILPs}

Previous work \cite{Crawford96,Aloul02a} has shown 
that breaking symmetries in CNF formulas
effectively prunes the search space and can lead to significant runtime
speedups. Breaking symmetries prevents symmetric images
of search paths from being searched, thus pruning the search tree.
The papers cited in this work all use variants
of the approach first described by Crawford et al. \cite{Crawford96}, which detects symmetries in 
a CNF formula using graph automorphism. The formula 
is expressed as an undirected graph such that the 
symmetry group of the graph is isomorphic to the symmetry
group of the CNF formula. Symmetries induce 
equivalence relations on the set of truth assignments of 
the CNF formula. All assignments in an equivalence class
result in the same truth value for the formula (satisfying or not). 
Therefore, it is only necessary to consider one assignment 
from each such class.

Techniques for symmetry breaking proposed in the literature
follow the following steps: (i) construction of a colored
graph from a CNF formula (ii) detection of symmetries in
the graph using graph automorphism  software (iii)
use of the detected symmetries to construct symmetry breaking
predicates (SBPs) that can be appended as additional
clauses to the CNF formula (iv) solution of the new CNF formula
thus created using a SAT solver. 
 Crawford's construction \cite{Crawford92} uses 3 colors for vertices,
one for positive literals, one for negative literals and a
third for clauses. Edges are added between literals in a clause
and the corresponding clause vertex, and between positive
and negative literal vertices for Boolean consistency. 
As an optimization, binary clauses (with just two literals)
are represented by adding an edge between the two involved
literals, so an extra vertex is not needed. This is useful because
the runtime of graph automorphism programs such as {\sc Nauty} \cite{Nauty}
 generally increases with the number of vertices in the graph. However,
with this optimization Boolean consistency is not enforced, since
binary clausal edges could be confused with Boolean
consistency edges between positive and negative literals
of the same variable. This may be improved by
representing binary clausal 
edges as {\em double edges} \cite{Crawford96}, 
thus distinguishing between the two edge types. 
However, {\sc Nauty} (and other graph automorphism programs) do not
support the uses of double edges, so this construction
is not very useful in practice. Furthermore, the cited constructions \cite{Crawford92,Crawford96}
 do not allow detection of {\em phase-shift} symmetries, when a variable's positive 
literal is mapped to its negative literal and vice versa, since
they color positive and negative literals differently.
Our previous work \cite{Aloul02a} improves upon these
constructions by giving positive and negative literal
vertices the same color, and allowing binary clauses {\em and}
Boolean consistency edges to be represented the same way,
i.e. a single edge between two literal vertices. Although this construction
may allow spurious symmetries - when clause edges are mapped
into consistency edges - this can occur only when a 
formula contains {\em circular chains of implications}
over a subset of its variables. For example, given
a subset of variables $x_1 \ldots x_n$, such a chain is a
collection of clauses $(y_1 \Rightarrow y_2)(y_2 \Rightarrow y_3)
\ldots (y_{n-1} \Rightarrow y_n)$, where each $y_i$
is a positive or negative literal of $x_i$. These circular
chains rarely occur in practice, and can be easily checked for.
Therefore, the efficient graph construction described above can be
used in most practical cases. 

Graph automorphisms are detected in Crawford's work \cite{Crawford96} as
well as our previous work \cite{Aloul02a} using the program {\sc Nauty} \cite{Nauty},
which is part of the GAP (Groups, Algebra and Programming) package. 
{\sc Nauty} accepts graphs in the GAP input format and returns a
list of {\em generators} for the automorphism group (the term
``generators'' is used in a mathematical sense, the symmetry group 
partitions the set of vertex permutations for the graph into equivalence
classes such that all permutations in the same class are equivalent.
{\sc Nauty} returns the set of generators for this symmetry group).
More recent work (\cite{Aloul03a,Aloul04}) uses the
automorphism program {\sc Saucy} \cite{Saucy}, which is more
efficient than {\sc Nauty} and can also process larger graphs with
more vertices. After generators of the symmetry group are detected,
symmetry breaking predicates are added to the instance
in a pre-processing step. Crawford et al. \cite{Crawford96} propose  
 the addition of SBPs that choose lexicographically smallest assignments (lex-leaders)
from each equivalence class. We refer to such SBPs as 
instance-dependent SBPs, since the symmetries
are first detected and then broken, and therefore
the exact number and nature of SBPs added always depends
on the connectivity of the graph itself. Although
detecting symmetries is non-trivial, using modern software
such as {\sc Nauty} and {\sc Saucy} the detection time is frequently
insignificant when compared with SAT-solving time. 
Crawford et. al. \cite{Crawford96} construct lex-leader 
SBPs for the entire symmetry group, using the group generators
returned by {\sc Nauty}. This type of symmetry breaking is
{\em complete}. However, the approach used by Aloul et al. in TCAD 2003 \cite{Aloul02a} shows that
{\em incomplete} symmetry breaking, which breaks      
symmetries only between generators, is often effective in practice
and much more efficient since it does not require the whole
group to be reconstructed. The SBP construction proposed in the cited work \cite{Aloul02a}
is {\em quadratic} in the number of problem variables, compared
with the earlier construction \cite{Crawford96}, which could run to exponential
size. This construction is further improved in the 2003 work by Aloul, Sakallah and Markov \cite{Aloul03a}, which
describes efficient, tautology-free SBP construction, whose
size is linear in the number of problem variables. 
Empirical results from both Crawford's work \cite{Crawford96} as well as the work in TCAD 2003 \cite{Aloul02a} show
that breaking symmetries produces large search speedups on
a number of CNF benchmark families, including pigeonhole
and Urquhart benchmarks, microprocessor verification,
FPGA routing and ASIC global routing benchmarks from the
VLSI domain.

Our work on symmetry breaking in SAT \cite{Aloul02a} has also been extended to  
  to optimization problems that include both CNF and PB 
constraints, and an objective function \cite{Aloul04}.
 As before, symmetries are detected
by reduction to graph automorphism. A PB formula for
an optimization problem is represented by an undirected graph.
Graph symmetries are detected using the graph automorphism tool
{\sc Saucy} \cite{Saucy}.
 Efficient symmetry breaking predicates \cite{Aloul03a} are appended to the formula as CNF clauses.
The empirical results for our work on symmetry breaking in 0-1 ILP \cite{Aloul04} show that the addition
of symmetry breaking predicates to PB formulas results
in considerable search speedups for the specialized 0-1
ILP solver PBS \cite{Aloul02b}. In this work, we use the above methodology 
  \cite{Aloul04} for detecting and breaking {\em instance-dependent}
symmetries in instances of graph coloring expressed as 0-1 ILP.
These instance-dependent SBPs are compared with a
number of instance-independent SBP constructions described
in the next section.

Detecting and breaking symmetries in application-derived
SAT instances amounts to a recovery of structure from
the original application. The loss of structure during problem 
reductions is one reason why reduction-based techniques are
often not competitive with domain-specific algorithms, 
and recent work on symmetry breaking is useful in this context.
Other types of structure include {\em clusters} \cite{Huang03,Aloul04b}. 
 Huang et al. \cite{Huang03} propose an algorithm that detects clusters in
SAT instances and uses them to produce variable orderings,
and these structure-aware orderings result in considerable empirical
improvements with the SAT solver zChaff \cite{Chaff}.

\subsection{Reducing Graph Coloring to 0-1 ILP}
We express 
an instance of the minimal graph coloring problem
as a 0-1 ILP optimization problem, consisting of 
(i) CNF and PB constraints that model the graph 
(ii) an objective function to minimize the number of colors used.

Consider a graph $G(V,E)$. Let $n = |V|$ be the number
of vertices in $G$, and $m = |E|$ be the number of edges. 
An instance of the $K-$coloring
problem for $G$ (i.e., can the vertices in $V$ be colored
 with  $K$ colors) 
is formulated as follows.
\begin{itemize}
\item
For each vertex $v_i$, $K$ {\em indicator variables} $x_{i,1}, \ldots ,x_{i,K}$,
denote possible color assignments to $v_i$.
Variable $x_{i,j}$ is set to 1 to indicate that vertex $v_i$ 
is colored with color $j$, and 0 otherwise 
\item For each vertex $v_i$,
a PB constraint of the form
$\sum_{j=1}^K x_{i,j} =1$ 
ensures that each vertex is colored with {\em exactly} one color.
\item Each edge $e_i$ in $E$ connects two vertices
$(v_a, v_b)$. For each edge $e_i$, we define 
CNF constraints of the form
$\bigwedge_{j=1}^{K} (\overline{x_{a,j}} \vee \overline{x_{b,j}})$ 
to  specify that no two vertices connected by an 
edge can be given the same color.
\item To track used colors,
we define $K$ new variables, $y_1, \ldots, y_K$. 
Variable $y_i$ is true {\em if and only if} at least one vertex 
uses color $i$. This
is expressed using the following CNF constraints:
$\bigwedge_{j=1}^K \left( y_j \Leftrightarrow \left( \bigvee_{i=1}^n x_{i,j} \right) \right)$.
\item The optimization objective is to minimize 
the number of $y_i$ variables set to true, i.e. MIN
$\sum_{i=1}^{K} y_i$
\end{itemize}

The total number of variables in the formula is $nK+K$.
The total number of constraints is computed as follows.
There are totally $n$ 0-1 ILP constraints (one per vertex)
to ensure that each vertex uses exactly one color.
For each edge, there are $K$ CNF clauses specifying that
the two vertices connected by that edge cannot have the same
color, giving a total of $mK$ CNF clauses. There are an additional
$nK$ CNF clauses ($K$ per vertex) for setting indicator
variables, and $K$ CNF clauses, one per color, to complete
the iff condition for indicator variables. This gives a total
of $K\cdot(m + n + 1)$ CNF clauses and $n$ 0-1 ILP
constraints, plus one objective function, in the converted
formula. For dense graphs, where $|E| \approx {|V|}^2$,
the resulting formula size is quadratic in the number of vertices
of the graph, but for sparser graphs it may be linear.
 A key observation is that instance-dependent symmetries in
graph coloring survive the above reduction to 0-1 ILP.
For instance-independent symmetries (i.e. permutations of
colors) this is easy to see, since the ordering of
colors can be changed without having any effect on
the formula and producing the same set of constraints.
For instance-dependent symmetries, consider two vertices
$v_a$ and $v_b$ that are symmetric to each other
and can be swapped in the original graph. Clearly, the constraints
that specify that $v_a$ and $v_b$ must use exactly
one color are interchangeable, as are the constraints
that determine color usage based on the colors assigned
to $v_a$ and $v_b$. It only remains to show that the 
{\em connectivity constraints} that control colors of vertices
adjacent to $v_a$  and $v_b$ are also symmetric. This is clear
from the fact that for every edge $E_i$ incident on $v_a$,
there must be a corresponding edge $E_j$ incident on $v_b$
for the two vertices to be symmetric ($E_i$ and $E_j$ can be the same
edge). Therefore, for the set of $K$ CNF clauses added to
the formula to represent $E_i$, there must be a symmetric
set of clauses added for $E_j$, and thus connectivity is preserved.

It is also clear that the 0-1 ILP formulation does not introduce spurious
symmetries, i.e. any symmetry in the formula is a symmetry in the
graph. A spurious symmetry arises when (i) variables of different
types can be mapped into each other, e.g. vertex color variables
are mapped to color usage indicator variables and (ii) variables
of the same type are mapped into each other when the corresponding
vertices are not actually symmetric. From the construction of
the 0-1 ILP formula, it is clear that all $K$ variables
per vertex that indicate a vertex's color can be permuted,
as can the $K$ color usage variables, since these all
appear in exactly the same constraints. This corresponds to
the instance-independent symmetry - colors in an instance of
graph coloring can be arbitrarily permuted. However, vertex
color variables appear in constraints restricting the number
of colors a vertex can use and also in constraints that
describe the connectivity of the graph, whereas 
color usage variables appear {\em only} in constraints
that specify when they are set. Therefore, the two types
of variables cannot map to one another. Since all constraints
regarding color and connectivity of a vertex are written using
{\em all} $K$ color variables for that vertex, these variables
are symmetric to each other only in groups of $K$, i.e.
if one such variable for a given vertex $v_1$ is symmetric to
a variable for another vertex $v_2$, then all $K$ variables 
for $v_1$ and $v_2$ are correspondingly symmetric. Additionally,
this symmetry between variables indicates a correspondence
of clauses in which they occur. This is only possible if
the vertices $v_1$ and $v_2$ are symmetric in terms of connectivity
(instance-dependent symmetry). Thus, both types of symmetries
are preserved during conversion to 0-1 ILP, and no
false symmetries are added. Therefore, we can apply known techniques
for symmetry detection in 0-1 ILP.


\section{Instance-Independent SBPs}
\label{sec:sbconst}

The question addressed in this work is whether 
{\em instance-independent} SBPs added during the reduction
can provide even greater speedups, possibly by accelerating
the detection of instance-dependent symmetries.
 To answer this question, we propose three provably correct SBP constructions
of varying strength, and one heuristic that is intended to break a small number of symmetries
 with minimal overhead. 
Each construction 
is implemented and empirical results are reported in Section \ref{sec:results}.

We use the following notation.
Consider an instance of the $K-$coloring problem,
which asks whether a graph $G(V,E)$ can be colored
using $\leq K$ colors and minimizes the number of colors.
Assume the colors are numbered $1 \ldots K$. We denote a valid 
color assignment by $(n_1, n_2, \ldots , n_K)$ 
where $n_i$ is the number of vertices colored with
color $i$, and $|V|=\sum_{i=1}^{K} n_i$.
Each $n_i$ in the color assignment denotes the cardinality of
the independent set colored with color $i$. We are not
concerned with the actual {\em composition} of the independent
sets here, since that is an instance-dependent issue. Instance-independent
symmetries are only the arbitrary permutations of colors between
different independent sets. 

\begin{figure*}
\begin{center}
\epsfysize=17cm
\epsffile{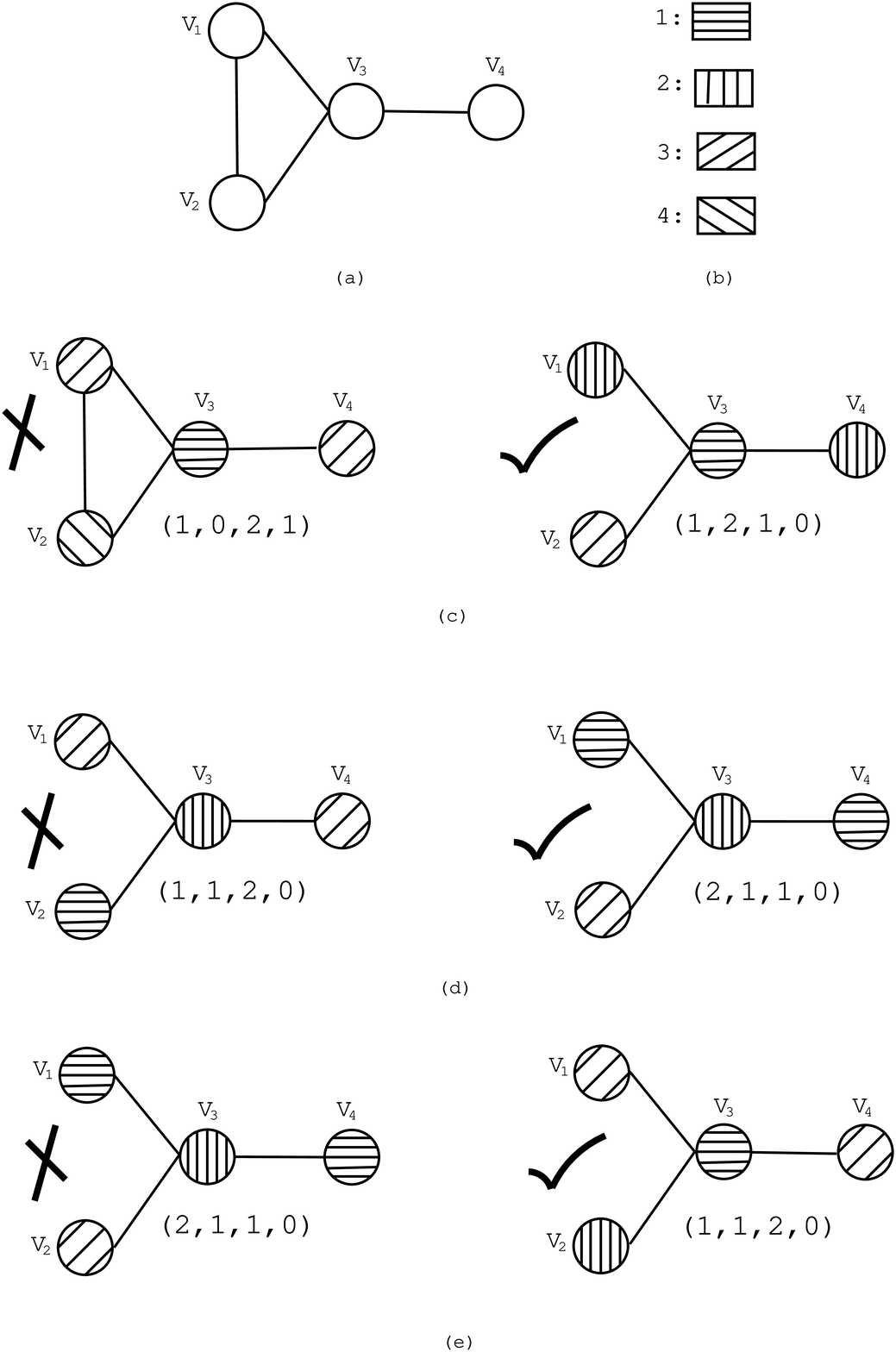}
\parbox{13.5cm} {
\caption{
    \label{fig:iisym}   
     {\bf Instance-independent symmetry breaking predicates (SBPs).
Part (a) shows the original graph with no vertices colored. Part (b)
shows the color key. Part (c) shows how null-color SBPs prevent
color 4 from being used. Part (d) shows how cardinality based SBPs 
assign colors in the order of independent set sizes, allowing fewer
assignments than null-color SBPs. Part (e) demonstrates 
how lowest-index ordered SBPs break symmetries that are undetected by other 
types of SBPs. 
}
}
}
\end{center}
\end{figure*}

The effects of each proposed construction 
 are illustrated using the example in Figure \ref{fig:iisym}.
The figure is an example of the 4-coloring problem on a graph with
four vertices. Part (a) of the figure shows the graph to be colored.
For visual clarity, part (b) shows color patterns corresponding to the
different color numbers. It is clear from the figure that the vertices
$V_1$, $V_2$ and $V_3$ form a clique, and must use different colors.
However, $V_4$ can be given the same color as either $V_1$ or $V_2$,
and therefore only 3 colors are needed for this instance. The instance
can be partitioned into independent sets in two ways:
$\{ \{V_1, V_4 \}, \{ V_2 \}, \{ V_3 \} \}$ and 
$\{ \{V_1 \}, \{ V_2, V_4 \}, \{ V_3 \} \}$. Our SBPs
do not actually address how the independent sets are composed, because
this is an instance-dependent issue. However, given any partition
of independent sets, colors can be arbitrarily permuted between
sets in the partition. The instance-independent SBPs proposed
here restrict this permutation. In the examples below, we assume
the first partition of independent sets i.e. 
$\{ \{V_1, V_4 \}, \{ V_2 \}, \{ V_3 \} \}$. Results are proved with
respect to the permutation of colors for this partition.

\subsection{Null-Color Elimination (NU)}
Consider a  $K-$coloring problem with colors $1 \ldots K$
for a graph $G(V,E)$.  
Assume that $G$ can be minimally colored with
 $K-1$ colors. Consider an optimal solution where 
color $i$ is not used: $(n_1, n_2, .. n_{i-1}, 0, n_{i+1}, \ldots ,n_K)$.
This assignment is equivalent to another assignment, 
 $$({n^\prime}_1, {n^\prime}_2, .. {n^\prime}_{j-1}, 0, {n^\prime}_{j+1} ... {n^\prime}_K)$$
where $i \neq j$ and ${n^\prime}_i=n_j$. 
For example, the assignment $(1,0,2,3)$ is equivalent
to $(1,3,2,0)$, $(0,1,2,3)$, $(1,2,0,3)$.
This is due to the existence of
{\em null} colors, which
create symmetries in an instance of
$K-$coloring because any
color can be swapped with a null color. Null colors are extraneous
because they are not actually required to color any vertices,
and so can be inserted anywhere in a solution, as seen above.
We propose a construction that enforces
an ordering on null colors: null colors
may appear only at the {\em end} of a color
assignment, after all non-null colors.
This is implemented by adding $K-1$ CNF constraints
of the form: $y_{k+1} \Rightarrow y_k$ for 
$1 \leq k \leq K-1$, to the original formulation.
In the example above, only one of the four 
symmetric assignments $(1,3,2,0)$ would be allowed
under this construction. Since our ILP formulation defines
and sets the $K$ indicator variables that track color usage, it
is extremely easy to enforce null color elimination as
described above. The SBPs require the addition of no extra variables
and only $K-1$ new CNF clauses.

We prove that the proposed construction is correct. 
Assume that under the original formulation, an
optimal solution  for graph $G(V,E)$
uses $m$ colors. Assume that this solution contains
null colors and non-null colors,
and with null-color elimination, there is a
{\em different} optimal solution that uses $m^\prime$
colors, where $m \neq m^\prime$. The only colors
used in this solution are $1 \ldots m^\prime$, since
null colors cannot occur before non-null colors.
Since our construction adds SBPs without
changing the original
constraints, any legal solution
that satisfies the SBPs will satisfy all constraints
in the original formulation. The solution to 
the original satisfies all constraints in the new formulation
except the SBPs. 
If $m < m^\prime$, we can re-order the solution so that all
null colors are placed last. This will satisfy all SBPs and use $m$ colors,
where  $m < m^\prime$, violating the assumption that the $m^\prime$-color 
solution was optimal. 
If  $m^\prime < m$, we already have a solution that satisfies all the original
constraints and uses fewer colors, which again
violates assumptions of optimality.

An illustration of the use of NU predicates for the example in Figure \ref{fig:iisym} (a) 
is shown in Figure \ref{fig:iisym} (c). The figure shows two valid 
minimal-color assignments to the graph vertices in the example. The assignment
on the left uses colors 1, 3 and 4, while the one on the right uses colors
1, 2 and 3. The assignments are symmetric but under NU predicates only
the right-hand side assignment is permissible.

\subsection{Cardinality-Based Color Ordering (CA)}
Null-color elimination 
is useful {\em only} in cases where null colors exist.
For a $K-$coloring problem where all colors are needed,
the construction breaks no symmetries.
Even when null colors exist, several
symmetries go undetected. In the first example from above,
null-color elimination permits six symmetric color assignments 
$(1,2,3,0)$, $(1,3,2,0)$, $(2,1,3,0)$, $(2,3,1,0)$ $(3,2,1,0)$ and $(3,1,2,0)$.
This is because restrictions are placed on null colors,
but the ordering of
non-null colors is unrestricted. A stronger 
construction would distinguish between the independent
sets themselves.
We propose an alternative construction,
which assigns colors based on the cardinality of independent
sets. This subsumes null-color elimination, since null colors can 
be viewed as coloring sets of cardinality 0.
The cardinality rule is implemented as follows:
the largest independent set is assigned the color 1,
the second-largest the color 2, etc. 
In the example above, {\em only} the assignment $(3,2,1,0)$ is valid.
This is enforced by adding $K-1$ PB constraints of the form:
$\sum_{i=1}^n x_{i,k} \geq \sum_{i=1}^n x_{i,k+1}$, where
$1 \leq k \leq K-1$. Again, this construction is fairly
simple to implement, requiring only $K-1$ additional constraints.
However, these are 0-1 ILP constraints with multiple variables,
unlike the simple CNF implication clauses between two variables used for
the NU predicates. Thus, there is some overhead for greater
completeness.

We prove the CA construction correct as follows.
Assume an optimal solution under this
construction uses $m < K$ colors: $(n_1,n_2, \ldots ,n_m)$,
where ($n_1 \geq n_2 \ldots \geq n_m$).
Colors $> m$ are not used on any vertex,
Assume there exists an optimal solution to the original formulation 
that uses $m^\prime$ colors: 
$({n^\prime}_1, {n^\prime}_2, \ldots ,{n^\prime}_{m^\prime})$, 
(where ${n^\prime}_1$, etc. are not arranged in descending
order). Without loss of generality, assume that $m^\prime < m$. 
We can sort the numbers ${n^\prime}_1, \ldots
 ,{n^\prime}_{m^\prime}$ and reassign colors in descending order. 
We would have a
solution with $m^\prime$ colors satisfying cardinality 
constraints. However, $m^\prime < m$, 
which is not possible if the $m-$color solution was
optimal. A similar argument applies when $m < m^\prime$. 

For the example from Figure \ref{fig:iisym} (a), { \em only}
the largest independent set under the partition we are considering,
i.e. $\{V_1, V_4\}$ can be given color 1. Therefore, the assignment
on the right of Figure \ref{fig:iisym} (c), which assigns the
largest set color 2 and is correct under
NU predicates, is incorrect under the CA construction. The left-hand
side of Figure \ref{fig:iisym} (d) shows another assignment that is 
correct under NU predicates but incorrect under CA predicates, since it assigns
the set $\{V_1, V_4\}$ color 3. A correct assignment, shown on the right-hand
side of Figure \ref{fig:iisym} (d), gives the largest set color 1 and since both
the other sets have one element each, they can each be assigned
either color 2 or color 3. Thus, several symmetric assignments which
survive NU predicates are prohibited under this construction.

\subsection{Lowest Index Color Ordering (LI)}
While more complete than NU predicates, CA predicates do not 
 break symmetries when different 
independent sets have the same cardinality. Consider a graph $G$ where $V=\{v_1, \ldots ,v_8\}$,
and an optimal solution, satisfying cardinality-based
ordering, that partitions $V$ into
4 independent sets: $S_1 = \{v_4, v_6, v_7\}$, 
$S_2 = \{v_1, v_5\}$,  $S_3 = \{v_3,v_8\}$,  $S_4 = \{v_2\}$.
A solution that assigns colors 2 and 3 to $S_2$ and $S_3$ is
symmetric to one that assigns colors 2 and 3 to $S_3$ and $S_2$.
Both are legal under cardinality-based ordering.
In order to completely break symmetries,
it is not adequate to distinguish between sets solely on the basis
of cardinality (unless no two sets have the same cardinality).
It is necessary to construct SBPs based on the actual 
{\em composition} of sets in a partition, which is unique.
However, the distinctions that we make on the basis of
composition are not to be confused with instance-dependent SBPs, 
since our construction is implemented {\em before} the symmetries
in an instance are known, and regardless of its actual
composition. The SBPs here specify broad guidelines for the
coloring of independent sets that are applicable to all graphs.
To improve upon cardinality-based ordering,
we propose a set of predicates to enforce the
{\em lowest-index ordering} (LI). 
 Consider all vertices with color $i$, and find the lowest index $j_i$ 
 among those. 
 We require that the lowest indices for each color 
 be ordered. This constraint
 can be enforced by adding inequalities for colors with 
 adjacent numbers.
 
Note that each color has a {\em unique} lowest-index vertex ---
otherwise some vertex would have to be colored with two colors.
In the above example, the only color assignment compatible with
the partitioning of vertices into independent sets is:
color 1 to $S_1$, 2 to $S_3$, 3 to $S_4$, 
and 4 to $S_2$.

 To evaluate the strength of this symmetry-breaking technique,
 consider an arbitrary coloring and a color permutation that remains
 a symmetry after the LI constraints are imposed. If any colors are
 permuted simultaneously on all vertices, this will permute the lowest indices
 for those colors.  Since all lowest indices are different, their ordering 
 is completely determined by the ordering of colors, and thus the color 
 permutation we chose must be the identity permutation. In other words,
 no instance-independent symmetries remain after symmetry-breaking with LI.
 

We implement lowest-index color ordering as follows.
For each vertex $v_i$, we declare
a new set of $K$ variables, $V_{i,1}, \ldots V_{i,K}$.
Variable $V_{i,k}$ being set implies that vertex
$v_i$ is the lowest-index vertex colored with color $k$.
This is enforced by the following CNF constraints:
$V_{i,k} \Rightarrow \left( \bigwedge_{j=1}^{i-1} \overline{V_{j,k}} \right)$.
Also, {\em exactly one} $V_{i,j}$ variable must be
true for every color used. Therefore, we add the constraints:
$y_k \Rightarrow \bigvee_{i=1}^n V_{i,k}$, where
$1 \leq k \leq K$, $y_k$ are the variables that indicate
color $k$ is used, and $n = |V|$ from Section \ref{sec:bkg}.
Finally, the following CNF clause is added for each $V_{i,k}$ to ensure
lowest-index ordering:
$V_{i,k} \Rightarrow \left( \bigvee_{j=i+1}^{n} V_{j,k-1} \right)$,
Since the LI ordering completely breaks symmetries between
independent sets, it subsumes earlier constructions.
However, it does come at an added cost. While the NU and CA
constructions required no new variables and only $K-1$ constraints,
the LI construction requires $nK$ new variables {\em and}
an additional $2nK$ CNF clauses, which is almost double the
size of the original formula.

The LI construction can be proved correct by the same means as the
CA construction. Given an optimal assignment of colors to
independent sets, we can sort the independent sets in order of
lowest-index vertex and assign colors from 1 to $K$ accordingly, 
without affecting correctness.

Figure \ref{fig:iisym} (e) illustrates the effect of LI SBPs
on the example in Figure \ref{fig:iisym} (a). The graph on the left,
which is shown as being correct for CA predicates in Figure \ref{fig:iisym}
(d) is incorrect under the LI construction, because the lowest-index vertex
with color 2 ($V_3$) does not have a higher index than the lowest-index
vertex with color 3, which is $V_2$.
 The graph on the right
shows the correct assignment, which under LI predicates is the
{\em only} permissible assignment for the partition $\{ \{ V_3 \},  \{ V_2 \}, \{V_1, V_4 \} \}$.

In addition to being very complex, LI predicates are 
so rigid that they obscure symmetries
of the original instance. For example, in Figure \ref{fig:iisym} (a),
it is easily seen that the vertices $V_1$ and $V_2$ are symmetric
and can be permuted with no effect on the resulting graph. This
symmetry is {\em instance-dependent} - it is decided by the way
$V_1$ and $V_2$ are connected. Without the addition of any SBPs,
it is apparent that under any legal coloring of the graph, the colors
given to $V_1$ and $V_2$ can be swapped regardless of how
$V_3$ and $V_4$ are colored. The NU predicates preserve this symmetry,
since they are only concerned with null colors which by definition could
not be used on $V_1$ and $V_2$. The CA predicates also preserve the symmetry
since $V_1$ and $V_2$ can be interchangeably used in any independent set, 
and swapping them between sets would not have any effect on the cardinality of
the sets. However, under the LI predicates, an independent set containing $V_1$ must {\em always}
be given a higher-numbered color than a set containing $V_2$, 
and the two cannot
be interchanged. If $V_1$ was given any color other than the highest
color in use, there would exist some independent set whose color index
was 1 greater than the color assigned to $V_1$, and for this set,
the lowest-index predicate would not be satisfied. 
Thus, LI predicates actually destroy any vertex
permutations in the graph. This is seen in our empirical results in 
Section \ref{sec:results}, where the addition of LI SBPs leaves
{\em no} symmetries in any of the benchmarks. This is unusual because
ordinarily benchmarks of reasonable size would contain at least some
vertex permutations.


\subsection{Selective Coloring (SC)}

It is noticeable that the ILP formulation and constraints 
can be very complex for more complete SBPs,
such as the LI predicates above, which introduce several
additional variables and clauses.
This raises the question of whether such a complex 
construction is actually counterproductive - it
may break symmetries, but 
require so much effort during search that the benefit 
of complete symmetry breaking is lost. To investigate this,
we also propose a simple ``heuristic'' construction
to break some symmetries 
between vertices while adding almost no 
 additional constraints. 
To impact as many vertices as possible,
we find the vertex $v_l$ with the largest degree
of all vertices in the graph. We then color $v_l$
with color 1. This is achieved by simply adding
the unary clause $x_{l,1}$. We search
$v_l$'s neighbors to find the vertex
$v_{l^\prime}$ with the highest degree out of all vertices
adjacent to $v_l$. We color $v_{l^\prime}$ with color 2,
by adding the unary clause $x_{l^\prime, 2}$. 
This construction has the effect of simplifying
color assignment for all vertices adjacent to $v_l$
and $v_{l^\prime}$. No vertex adjacent to $v_l$
can be colored color 1, and no vertex adjacent
to $v_{l^\prime}$ can be colored color 2. Moreover,
all vertices in an independent set with $v_l$
($v_{l^\prime}$) {\em must} be colored color 1 (color 2).
If $v_l$ and $v_{l^\prime}$ have sufficiently large degree,
this construction can restrict many vertex assignments.
An even stronger construction would be to find a triangular
clique and fix colors for all three vertices in it;
however, clique finding is complicated and some
graphs may not possess any such cliques.
We refer to this construction as {\em selective coloring}.

The extent to which selective coloring breaks symmetries
is instance-dependent. It fails to completely break symmetries
for almost all graphs. However, it is a simple
construction, adding just two constraints as 
unary clauses. These are easily resolved
in pre-processing by most SAT solvers,
so any symmetry breaking achieved by this construction
has virtually no overhead.

We note that all instance-independent predicates defined here are only concerned with 
symmetries between colors, which exist in any instance of graph coloring. However, additional instance-independent symmetries 
may be introduced during the reduction to graph coloring for certain applications. For example, in the radio
 frequency assignment application from Section \ref{sec:bkg}, adding all possible bipartite edges between cliques
 for adjacent regions will result in symmetries between vertices in these cliques. Additional predicates can be
added to instances from this application to break these symmetries. 

\section{Empirical Results}
\label{sec:results}

This section describes our experimental setup, empirical results, and performance
compared with related work.

\subsection{Experimental Setup}

We used 20 medium-sized instances from the DIMACS 
graph coloring benchmark suite.  We briefly describe each
family of benchmarks used below. 

\begin{itemize}
\item {\bf Random graphs. } Benchmarks with randomly created
connections between vertices, named {\tt DSJ}

\item {\bf Book graphs. } Edges represent interaction between 
characters in a book. There are four such benchmarks: 
{\tt anna, david, huck, jean} 

\item {\bf Mileage graphs. } These represent distances between
cities on a map, and are named {\tt miles} 

\item {\bf Football game graphs. } Indicate relationships
between teams that must play each other in 
college football games. In the tables these are referred to as
{\tt games}

\item {\bf $n-$queens graphs. } Instances of the $n-$queens problem,
named {\tt queen}

\item {\bf Register allocation graphs. } Represent the register
allocation problem for different systems. We use two families
in this work, named {\tt mulsol, zeroin} 

\item {\bf Mycielski graphs. } Instances of triangle-free graphs
 based on the Mycielski \cite{Myciel55} transformation, called {\tt myciel}
\end{itemize}

Table \ref{tab:bench} gives the name, size (number of vertices and edges) 
and the chromatic number for each benchmark.
We use a maximum value of $K=20$ for $K-$coloring.
For benchmarks with chromatic number $>20$, 
we do not report the chromatic number.
 
%
Our problem formulation with a fixed $K$ is application-driven. Indeed,
in many domains it is only useful to find the exact chromatic number 
when it is below a well-known threshold. For example, in graph coloring
instances from register allocation, there cannot be more colors than 
processor registers. PC processors often have 32 registers, and high-end
CPUs may have more. However, realistic graphs are relatively sparse and have 
low chromatic numbers.
On the other hand, processors embedded in cellular phones, automobiles
and point-of-sale terminals may have very few registers, leading to 
tighter constraints on acceptable chromatic numbers. 
The value $K=20$ used in our experiments is in no way special,
but the results achieved with it are representative of other results. 
 Also, while we apply the $K=20$ bound to all instances here to study trends,
 more reasonable bounds can be determined on a per-instance basis using the following simple
 procedure.
\begin{enumerate}
\item Apply any heuristic for min-coloring
       to determine a feasible upper bound
\item If the value is relatively small,
       perform linear search by incrementally
       tightening the color constraint,
       otherwise perform binary search
\end{enumerate}

 

\begin{table}[t]
\begin{center}
\footnotesize
\begin{tabular}{|l||c|c|c|}
\hline
Instance & \#V & \#E & $K$ \\ \hline			
anna &		138 &	986 &	11 \\
david &		87 &	812 &	11 \\
DSJC125.1 &	125 &	1472 &	5 \\
DSJC125.9 &	125 &	13922	& $>$ 20 \\
games120 &	120 &	1276 &	9 \\
huck	&	74 &	602 &	11 \\
jean	&	80&	508 &	10 \\
miles250&	128 &	774 &	8 \\
mulsol.i.2&	188&	3885&	$>$20 \\
mulsol.i.4&	185&	3946&	$>$20 \\
myciel3	&	11&	20&	4 \\
myciel4	&	23&	71&	5 \\
myciel5	&	47&	236 &	6 \\
queen5\_5&	25&	320 &	5 \\
queen6\_6 &	36 &	580 &	7 \\
queen7\_7 &	49 &	952 &	7 \\
queen8\_12 &	96 &	2736 &	12 \\
zeroin.i.1&	211&	4100&	$>$20 \\
zeroin.i.2 & 211 &	3541&	$>$20 \\
zeroin.i.3 &	206&	3540&	$>$20 \\ \hline
\end{tabular}

\parbox{3in}
{

\caption{ \label{tab:bench} \small \bf
DIMACS graph coloring benchmarks
}
}
\end{center} 
\vspace{-2mm}
\end{table}

Benchmark graphs are transformed into instances of 0-1 ILP using the
conversion described in Section \ref{sec:bkg}.
To solve instances of 0-1 ILP, we used the academic 0-1 ILP
solvers PBS \cite{Aloul02b}, Galena \cite{Chai03}, and Pueblo
\cite{Pueblo}, and also the commercial ILP solver CPLEX version 7.0.
Pueblo is more recent than PBS and Galena, and incorporates Pseudo-Boolean 
(PB) learning based on ILP
cutting-plane techniques. We use a later version of PBS, PBS II,
that enhances the original PBS algorithms \cite{Aloul02b} 
with learning techniques from the Pueblo solver \cite{Pueblo}.
We do not include the results with the original version of PBS
that are reported in \cite{Ramani04}, since it has been retired by the newer version. 
However, in the Appendix we report detailed results for 
$n-$ queens instances using the older version of PBS
along with results for the other solvers, for the sake of a more detailed study. 
PBS II is implemented in C++ and compiled using g++. 
Galena and Pueblo binaries were provided by the authors.
PBS was run using the variable state independent
decaying sum (VSIDS) decision heuristic 
option \cite{Chaff}. Galena was run using its default options of
linear search with cardinality reduction (CARD) learning.
All experiments are run on Sun-Blade-1000 workstations
with 2GB RAM, CPUs clocked at 750MHz and the Solaris operating system.
Time-out limits for all solvers are set at 1000 seconds.
 \begin{table}[t]
\begin{center}
\footnotesize
\begin{tabular}{|l||c|c|c||c|c|c|}
\hline
SBP & \multicolumn{3}{c||}{CNF Stats}  & \multicolumn{3}{c|}{Sym. Stats (SAUCY)}  \\ \cline{2-7}			
Type &	\#V & \#CL & \# PB & \#S & \#G & Time \\ \hline \hline
no SBPs & 437K & 777505 & 3193 & 1.1e+168 & 994 & 185 \\ \hline
NU & 437K & 777885 & 3193 & 5.0e+149 & 614 & 49 \\
CA & 437K & 777505 & 3630 & 5.0e+149 & 614 & 49 \\
LI & 870K & 4019980 & 3193 & 2.0e+01 & 0 & 84\\
SC & 437K & 777545 & 3193 & 3.0e+164 & 941 & 167 \\
NU+SC  & 437K & 777925 & 3193 & 5.0e+148 & 597 & 47 \\ \hline

\end{tabular}
\parbox{4in}
{
\caption{ \label{tab:symresult} \small \bf
CNF formula sizes, symmetry detection results and runtimes, 
totaled for 20 benchmarks from Table \ref{tab:bench}, with $K=20$.
NU = null-color elimination; CA = cardinality-based; LI = lowest-index; SC = 
selective coloring. For the LI SBPs, one instance of the
``do-nothing'' symmetry is counted in each case, giving 
a total of 20 symmetries and 0 generators. {\sc Saucy} is run on an Intel
Xeon dual processor at 2 GHz running RedHat Linux 9.0.
} 
}
\end{center}
\end{table}

We use the symmetry breaking flow first proposed in our earlier work \cite{Aloul04} 
to detect and break symmetries
in our original ILP formulation from Section \ref{sec:bkg}.
This flow uses the tool {\sc Shatter} \cite{Aloul03a}, 
which uses the {\sc Saucy} \cite{Saucy}
graph automorphism program and the efficient SBP
construction from \cite{Aloul03a}.
We also check for unbroken symmetries in formulations produced
by each of the instance-independent constructions described 
in Section \ref{sec:sbconst}. Our runtimes for symmetry detection and for
 solving the reduced 0-1 ILP problems are reported in the next section.

\subsection{Runtimes for Symmetry Detection and 0-1 ILP Solving}
Table \ref{tab:symresult}
shows symmetry detection results and runtimes. The numbers reported
in the table are sums of individual results for all 20 benchmarks
used. We report statistics as sums because reporting results for all
of SBPs on all benchmarks would be space-consuming,
and would also not illustrate trends as clearly. This work is concerned
with characterizing the broad impact of symmetry breaking. However,
we show detailed results for the {\tt queens} instances in the Appendix.

The first column in the table indicates the type of construction:
we use {\bf no SBPs} for the basic formulation, {\bf NU}
for null-color elimination, {\bf CA} for 
cardinality-based ordering, {\bf LI} for lowest-index ordering,
and {\bf SC} for selective coloring (the last row shows
NU and SC in combination). The next three columns show the number
of variables, CNF clauses, and PB constraints in the problems.
The last three columns show the number of symmetries, number of
symmetry generators, and symmetry detection runtimes for {\sc Saucy}.
Henceforth, we will refer to instance-dependent SBPs 
as {\em external}, because they are added to an instance
{\em after} symmetries are detected and are not part of the problem 
formulation. The top row is separated from the bottom 5 rows 
because it represents statistics without instance-independent
SBPs. We observe that adding instance-independent SBPs during
problem formulation does cut down the symmetry detection
runtime considerably. {\sc Saucy} has a total runtime of 185 seconds when
no instance-independent SBPs are added, but its runtimes with
NU, CA, LI and NU + SC constructions are much smaller. Only the SC construction
has a comparable runtime because it is a heuristic and breaks very 
few symmetries. The columns showing numbers of symmetries and generators
support this observation: the NU, CA, LI and NU + SC constructions
all have far fewer symmetries than the top row, but the SC construction
has almost the same number. For these benchmarks, the LI construction, 
breaks {\em all} symmetries, even instance-dependent vertex permutations
that may exist in a graph. {\sc Saucy} reports finding no symmetries for this 
construction (except one instance of the do-nothing symmetry for each graph, 
 which is trivial).
However, {\sc Saucy} runtimes for this
construction are larger than for the NU, CA and NU + SC constructions
(85 seconds to approximately 49 seconds) even though there are no
symmetries in the instances after LI predicates are added. A likely
reason for this is the sharp increase in instance size caused 
by the LI construction. In general, the SC construction has very little
effect on the number of symmetries - when used by itself, it leaves
most symmetries intact, and when used with the NU construction, 
the improvement over the NU construction alone is very small.
 
\begin{table*}[!t]
\tiny
\begin{center}
\begin{tabular}{|l||c|c|c|c|c|c|c|c|c|c|c|c|c|c|c|c|}
\hline
 SBP & \multicolumn{4}{c|}{PBS II, PB Learning} & \multicolumn{4}{c|}{CPLEX} & \multicolumn{4}{c|}{Galena} & \multicolumn{4}{c|}{Pueblo} \\ \cline{2-17}
Type &  \multicolumn{2}{c|}{Orig.} & \multicolumn{2}{c|}{w/i.-d. SBPs} & \multicolumn{2}{c|}{Orig.} & \multicolumn{2}{c|}{w/i.-d. SBPs} & \multicolumn{2}{c|}{Orig.} & \multicolumn{2}{c|}{w/i.-d. SBPs} & \multicolumn{2}{c|}{Orig.} & \multicolumn{2}{c|}{w/i.-d. SBPs} \\ \cline{2-17}	
 & Tm. &  \#S & Tm. & \#S & Tm. & \#S & Tm. & \#S &
Tm. & \#S & Tm. & \#S & Tm. & \#S & Tm. & \#S  \\ \hline \hline
no SBPs &  17K & 3 & 4.2K & 16 & 6.3K & 14 & 13K & 7 & 1.7K & 2 & 3K & 17 & 18K & 3 & {\bf 1.6K} & {\bf 19}  \\ \hline
NU & 8.2K & 13 & 7.5K & 13 &  5.9K & 15 & {\bf 6.5K} & {\bf 15} & 8.3K  & 11 & 6.7K & 11 & 9.1K & 12 & 8.3K & 13  \\
CA & 13K & 6 & 12K & 8 & 11K & 11 & 11K & 10 &  19K & 1  & 17K & 3 & 9K & 12 & 10K & 12 \\	
LI  & 15K & 6 & 15K & 6 & 16K & 4 & 16K & 4 & 15K & 5 & 15K & 5 & 16K & 5 & 16K & 5 \\		
SC  & 14K & 6 & {\bf 65} & {\bf 20} & 5.3K & 15 & 12K & 8 & 16K & 4 & {\bf 94.4} & {\bf 20} & 15k & 5 & 2.1K & 18   \\		
NU+SC & {\bf 6.9K} & {\bf 14} & 6.8K & 14 & {\bf 4.5K} & {\bf 16} & 6.4K & 14 & {\bf 6.1K} & {\bf 14} & 6.1K & 14 & {\bf 7.3K} & {\bf 13} & 7.1K & 13 \\ \hline
\end{tabular}
\parbox{5.8in}
{
\caption{ \label{tab:main} \small \bf
Runtimes and number of solutions found before and after 
SBPs are added for all constructions using 
PBS II (with PB learning), CPLEX, Galena and Pueblo;
all experiments are run on SunBlade 1000 workstations.
Timeouts for all solvers were set at 1000s. The maximum
color limit is set at 20, instances with $k > 20$ are unsatisfiable 
under these formulations. This is not a comparison of 
solvers.  We solve ILP formulations with 
equal optimal values using different solvers to weed out solver-specific 
issues. Best results for a given solver are shown in boldface. In the entries, 
K denotes multiples of 1000s seconds rounded to the nearest integer.
} 
}
\end{center}
\end{table*}

Table \ref{tab:main} shows the effect of symmetry breaking on runtimes
 of PBS II \cite{Aloul02b}, CPLEX \cite{Cplex}, Galena \cite{Chai03} and Pueblo \cite{Pueblo}. The first column in the table specifies 
the construction type, followed by the total runtime for each solver (with and without the addition of instance-independent SBPs) 
 and the number of instances solved for the construction.
 For each solver, the best performance among all configurations (largest
number of instances solved and corresponding runtime)  is boldfaced.
Results are given first for 
the new version of PBS, PBS II based on \cite{Pueblo}, followed by CPLEX, 
Galena and Pueblo. Runtimes for the older version of PBS can be obtained from our earlier work \cite{Ramani04}. 
 To compare performance of an individual solver
for different constructions, observe the runtime and solution entries for
different rows in the same column, and to compare performance for different
solvers on the same constructions, observe numbers for the same row across
all columns.
We observe the following trends.

\begin{enumerate}
\item  All benchmarks possess a large number of symmetries.  Different
instance-independent SBPs achieve varying degrees of completeness: 
the lowest-index ordering (LI) breaks all symmetries in the benchmarks used, while the selective coloring (SC) SBP breaks the fewest symmetries. {\sc Saucy} runtimes
for residual symmetry detection after the addition of instance-independent 
SBPs are highest for the no SBPs construction and the SC construction,
since they possess the largest numbers of symmetries

\item For the case where no SBPs of any kind are added,
CPLEX performs well, solving 14 out of 
20 instances within the time limit. However, 
PBS II, Galena and Pueblo perform poorly - Galena solves only 2 instances
and PBS II and Pueblo each solve 3

\item  PBS II, Galena and Pueblo benefit considerably from instance-dependent
symmetry breaking. When instance-dependent SBPs
are used without any of the instance-independent
constructions we propose, PBS II solves 16 instances within
the time limit, while Galena and Pueblo solve 17 and 19 instances
respectively. However, CPLEX is hampered by the addition of 
instance-dependent SBPs, and solves only 7 instances in this case

\item Adding {\em only} instance-independent SBPs improves
performance for all specialized 0-1 ILP solvers over the no-SBP version. 
The best performance for PBS II, Galena and Pueblo is seen
for the NU + SC construction - PBS II and Galena solve 14 instances, and Pueblo
solves 13. For CPLEX, the NU + SC
construction shows marginal improvement over the no-SBPs case
(16 instances are solved), but the more complex constructions,
CA and LI, actually undermine performance - CPLEX solves only 4 instances
with the LI construction. In general, complex SBP constructions
perform much worse than simple ones. 
PBS II, Pueblo and Galena also perform poorly with the CA and LI constructions -
Galena solves only 1 instance with the CA construction with no help from
instance-dependent SBPs, and very few instances are solved with the LI construction
for any solver

\item Adding instance-independent SBPs alone does not
solve as many instances as adding instance-dependent SBPs to
the 
SBP-free formulation. 
The best performance seen
with instance-independent SBPs is 14 instances solved, by
Galena and PBS II, and 16 instances solved by CPLEX, with
the NU + SC construction. When instance-dependent SBPs
are added PBS II and Galena solve all 20 instances with the SC construction.
The CA and LI constructions leave very few (or none at all) 
symmetries to be broken by instance-dependent SBPs. Consequently,
there is almost no difference in results with and without instance-dependent
SBPs for these constructions. However, they do not achieve the
same performance improvements as instance-dependent SBPs, due
to their size and complexity

\item Using instance-dependent SBPs in conjunction with the SC construction
is useful. With this combination, PBS II and Galena solve
all 20 instances within the time limit, and Pueblo solves 18. 
Runtime is also considerably improved for PBS II and Galena -- 
PBS II solves all 20 instances in a total of 65 seconds, and Galena in
94.4 seconds. The best overall performance,
in terms of number of solutions and runtime, is seen with this combination.
In general, however, the SC construction is not dominant on its own.
Results for the SC construction alone are very similar to results with
no SBPs, and results for the NU + SC combination are very similar to
those achieved by using only NU SBPs. The SC construction is effective
at ``boosting'' the performance of other constructions

\item The three specialized 0-1 ILP solvers - PBS II, Galena and Pueblo,
exhibit the same performance trends with respect to the constructions used,
and their performances are all comparable, in terms of both the number of solutions
found and runtime
This indicates that the variations in performance are due to
the different SBPs, not due to differing solver implementations.
All solvers are independent implementations based on
the same algorithmic framework (the Davis-Logemann-Loveland 
backtrack search procedure), but PBS II and Galena also have
learning capabilities
\item Adding instance-dependent SBPs to any construction usually adversely
affects the performance of CPLEX. This has been previously noted in other work \cite{Aloul04}.  Since the CPLEX
algorithms and implementation are not available in the public domain,
it is difficult to account for this effect. However, PBS and Galena
with symmetry breaking significantly outperform CPLEX without symmetry breaking

\item We report results as the sum of runtimes for all instances
  to illustrate trends. On a per-instance basis, the same trends
  are displayed. For example, for the no-SBPs case in the top row,
  PBS II solves 3 instances and Galena solves 2, but the two instances
  solved by Galena are among those solved by PBS II. 
  In general, the same instances tend to be ``easy'' or ``difficult'' for the 0-1 ILP
  solvers, although CPLEX behaves differently. An example of this behavior
  for the {\tt queens} family of instances is illustrated in the 
  Appendix

\end{enumerate}
\begin{table*}[!t]
\tiny
\begin{center}
\begin{tabular}{|l||c|c|c|c|c|c|c|c|c|c|c|c|c|c|c|c|}
\hline
 SBP & \multicolumn{4}{c|}{PBS II, PB Learning} & \multicolumn{4}{c|}{CPLEX} & \multicolumn{4}{c|}{Galena} & \multicolumn{4}{c|}{Pueblo} \\ \cline{2-17}
Type &  \multicolumn{2}{c|}{Orig.} & \multicolumn{2}{c|}{w/i.-d. SBPs} & \multicolumn{2}{c|}{Orig.} & \multicolumn{2}{c|}{w/i.-d. SBPs} & \multicolumn{2}{c|}{Orig.} & \multicolumn{2}{c|}{w/i.-d. SBPs} & \multicolumn{2}{c|}{Orig.} & \multicolumn{2}{c|}{w/i.-d. SBPs} \\ \cline{2-17}	
 & Tm. &  \#S & Tm. & \#S & Tm. & \#S & Tm. & \#S &
Tm. & \#S & Tm. & \#S & Tm. & \#S & Tm. & \#S  \\ \hline \hline
No SBPs &  18K & 2 & 6.2K  & 14	 & 11K	& 9 & {\bf 8.2K} & {\bf 12} & 19K & 1 & 9.1K & 11 &	19K & 1	& 7.5K & 13 \\ \hline
NU & 9.2K & 12 & 7.9K & 13 & 11K & 9 & 12K & 8 & 10K & 10 & 7.6K & 13 &	11K & 11 & 9.5K & 11 \\
CA & 13K & 7 & 13K & 9	& 13K & 9 & 14K	& 8 & 19K & 1 & 17K & 4	& 11K & 11 & 13K & 8 \\
LI & 15K & 5 & 15K & 5	& 19K & 2 & 19K & 2 & 16K & 5 & 16K & 5	& 17K & 3 & 17K & 3 \\
SC & 15K & 5 & {\bf 5.3K} & {\bf 15} & 10K & 10 & 12K & 9 & 16K & 4 & {\bf 5.3K}  &  {\bf 15} & 16K & 4 & {\bf 6.0K} & {\bf 15} \\
NU + SC & {\bf 7.1K} & {\bf 13} & 7.0K & 13 & {\bf 9.7K} & {\bf 11} & 9.9K & 11 & {\bf 9.2K} & {\bf 12} & 6.9K & 14 & {\bf 8.0K} & {\bf 13} & 7.4K & 13 \\ \hline
\end{tabular}
\parbox{5.8in}
{
\caption{ \label{tab:main2} \small \bf
Total runtimes and number of solutions found before and after 
SBPs are added for all constructions using 
PBS II (with PB learning), CPLEX, Galena and Pueblo.
The experimental setup is the same as that used in Table \ref{tab:main}
but with a color limit of $K = 30$. Best results for a solver
are boldfaced. Fewer instances are solved than in Table \ref{tab:main}
because the higher color limit results in larger and potentially more
difficult instances.
} 
}
\end{center}
\end{table*}
Overall, the results suggest that for graph coloring,
adding instance-independent SBPs alone is not competitive with the use of 
instance-dependent SBPs alone. The best results
are achieved using a combination of both types,
and even here, the instance-independent SBPs used are 
the most simple variety. 
This is true even when symmetry detection runtimes are taken into
consideration. 
We attribute this result to the complexity of
instance-independent SBPs we use, and also to the fact that improvements
in graph automorphism software \cite{Saucy} have greatly reduced the
overhead of detecting symmetries by reduction to graph automorphism.
Previously, for static approaches that require symmetries to be detected
and broken in advance, the task of symmetry detection was often
a bottleneck that could actually take longer than the search itself.
With this bottleneck removed, the advantages of static symmetry breaking - 
simple predicates that address specific symmetries rather than complex 
constructions that alter the problem specification considerably - are more
clearly illustrated. Even among instance-independent predicates, simple
constructions are more effective than complex ones. 
As we have noted in Section \ref{sec:sbconst}, simple
constructions like NU and SC add very few additional constraints
and do not alter the original problem greatly. However, the CA and LI
constructions add many more constraints, which may confuse the 
specialized 0-1 ILP solvers.

 It is important to note that color permutations, while instance-independent,
 do appear at the instance-specific level. Thus, the symmetries targeted by instance-independent
 predicates are a subset of those targeted by instance-dependent predicates. Our instance-independent
constructions are not intended to cover a different set of symmetries, but rather to break some of the same
 symmetries during problem formulation, thus reducing or eliminating the overhead of any instance-dependent methods
 that may follow. The fact that this strategy is not successful suggests that, for the same set of symmetries,
 the instance-dependent predicates we use are more efficient and easier for solvers to tackle.

To verify our claims about performance trends, we show
results for an additional set of experiments with increased
color limit $K = 30$ in Table \ref{tab:main2}. The instances
are re-formulated with $K = 30$ and with different SBP constructions.
This experiment is intended to verify trends from the $K = 20$
case, and to investigate whether instances with chromatic
number $> 20$, that are unsatisfiable in the first case,
can be colored with $\leq 30$ colors. Results from Table \ref{tab:main2}
validate our observations from Table \ref{tab:main} -- the best results 
for PBS II, Galena and Pueblo are again achieved with the NU + SC
(with no instance-dependent SBPs) and SC (with instance-dependent
SBPs) constructions. However, with this formulation {\em fewer}
instances are solved than for the $K = 20$ case, possibly because
the $K = 30$ limit results in larger instances. Also, for instances
whose chromatic number is much closer to 30 than 20, it may be harder
to prove optimality, whereas proving unsatisfiability for the $K = 20$ 
experiments may be simpler.

\subsection{Comparison with Related Work}
\label{subsec:relwork}

Here, we discuss the empirical performance of our approach when compared with 
related work \cite{Coudert97,Benhamou04}. We note that both cited works
 describe algorithms specifically developed for graph coloring, and the
search procedures cannot be used to solve other problems. Our approach, on 
the other hand, solves hard problems by reduction to generic problems such as 
SAT or 0-1 ILP, and this work on graph coloring can be viewed as a case study. 
 Consequently, we use problem-specific knowledge only during the actual problem 
 formulation (instance-independent SBPs are also added during reduction), but not during
search itself. This may be useful for applications where problem-specific solvers cannot be
developed or acquired due to limited resources. Our goal is to determine whether 
symmetry breaking can improve the performance of reduction-based methods, which 
are traditionally not competitive with problem-specific methods. Thus, while 
our techniques may not be superior to all problem-specific solvers on all instances, 
we hope to show  reasonably strong performance over a broad spectrum of instances.

Common data points between our work and Coudert's \cite{Coudert97}   
include instances of {\tt queens}, {\tt myciel} and {\tt DSCJ125.1}.
 Referring to our detailed results for {\tt queens} instances in the 
 Appendix, we note that our runtimes are competitive with those of Coudert's algorithm - 
 for example, on {\tt queen5$\_$5}, both algorithms have a runtime of 0.01s.
On larger instances, however, our runtimes are somewhat slower. On the {\tt myciel} instances,
 we obtain the best results with the Pueblo solver and the SC predicates, with runtimes
 of 0.01, 0.06, and 1.80s on {\tt myciel}3, 4, and 5, compared with 0.01, 0.02 and 4.17
 for Coudert's algorithm. Therefore, it appears that our approach
 is competitive on these common data points.
 Moreover, other studies \cite{KirovskiP98} have observed that 
 Coudert's work does not provide results for several hard real-world 
problem classes, particularly those where modeling results in dense graphs. 
Our work is more general, and cannot be biased to favor certain types of graphs.

The algorithm described by Benhamou \cite{Benhamou04} shows very competitive
runtimes on a number of DIMACS benchmarks, particularly instances of register 
allocation. For example, the {\tt DSJC125.1} instance is solved by Benhamou's algorithm 
in 0.01 seconds, while the best time achieved by us is 1.12 seconds, using the Pueblo solver
 with only instance-dependent SBPs. However, we note that Benhamou's algorithm determines the upper
 limit for the chromatic number $K$  using more instance-specific knowledge, for
example, for {\tt DSCJ125.1}, it is set at $K=5$. We solve all instances 
with $K=20$, which
may be too large a limit in some cases. The value of $K$ affects the 
size of the resulting
0-1 ILP reductions and SBPs, which is likely to affect runtime.
 We also note that the DIMACS benchmarks used in the cited work \cite{Benhamou04} are primarily 
 register allocation and randomly generated instances, whereas we achieve 
 reasonably good performance on a wide variety of benchmark applications. 
 Moreover, Benhamou's approach relies on modeling graph 
coloring as a not-equals CSP, which does not bode well for generality. 
 Many CSPs cannot be modeled using only not-equals constraints. Additionally,  
 the symmetry detection, breaking and search procedures described in that work  
 are specific to graph coloring, whereas our work can be extended to several 
 other problems, only requiring a reduction to SAT/0-1 ILP.  

\section{Conclusions}
\label{sec:conclude}
   Our work shows that problem reduction to 0-1 ILP is a viable
 method for optimally solving combinatorial problems without
 investing in specialized solvers. This approach is likely to
 be even more successful as the efficiency of 0-1 ILP solvers improves 
 in the future, and as they are able to better handle problem structure.
 In particular, problem reductions may produce highly-structured
 instances making the ability to automatically detect and exploit
 structure very important. In the case of graph coloring 
 we demonstrate that a generic, publicly-available symmetry breaking flow from
 our earlier work \cite{Aloul04} significantly improves empirical results in
 conjunction with the academic 0-1 ILP solvers PBS II, a new version of the solver PBS      
 \cite{Aloul02b}, Galena \cite{Chai03} and Pueblo \cite{Pueblo}.
 All specialized 0-1 ILP solvers significantly outperform the commercial
generic ILP solver CPLEX 7.0 when symmetry-breaking is used.  The
performance of CPLEX actually deteriorates when SBPs are added, and on the
original instances with no SBPs, CPLEX is able to solve more instances than the
0-1 ILP solvers. However, the best performance overall is obtained with the 0-1
ILP solvers on instances with SBPs added. Although our techniques are tested
on standard DIMACS benchmarks instances, we note that the symmetry-breaking
flow described here can be applied to graph coloring instances from any
application.

   We are particularly interested in comparing strategies for breaking 
 symmetries that are present in every ILP instance produced
by problem reduction (instance-independent symmetries). Such symmetries
  may be known even before the first instances of the original
 problem are delivered (i.e., symmetries may be detected at the specification
 level), and one has the option to use them during problem reduction. 
Intuitively,  this may prevent discovering these symmetries in every 
 instance and thus improve the overall CPU time. To this end, we propose
four constructions for instance-independent symmetry breaking predicates
(SBPs). These constructions vary in terms of strength and completeness.
Our goal in experiments was to compare the performance of the four 
instance-independent SBP constructions relative to each other, as well as to
assess their performance when compared with instance-dependent SBPs.
Instance-independent SBPs have the advantage of not requiring the additional
step of symmetry detection, since they are part of the problem specification.
Additionally, they are designed with more information about the problem itself, and their effect on solutions is clear - for example, we know that null-color elimination will force all the lower-numbered colors to be used in a solution.
Instance-dependent SBPs are detected and added automatically on the 0-1 ILP
reduction of an instance without any understanding of their significance.
On the other hand, instance-dependent constructions are less complex and result
in more compact predicates. Our empirical data indicate that simplicity
of construction is a more powerful factor in determining performance - 
 instance-dependent SBPs consistently outperform 
instance-independent SBPs, and the most complete and complex 
instance-independent constructions (LI) are actually the weakest in 
performance. It is clear
from our results that symmetry breaking itself {\em is}
useful in graph coloring: adding instance-dependent SBPs always
speeds up search over the no-SBPs case. It is likely that instance-independent
SBPs are less successful due to their complex construction. 
Simpler instance-independent
constructions (NU, SC) outperform the more complex ones (CA, LI).
It is well known that the syntactic structure of CNF
and PB constraints may dramatically affect the efficiency of SAT and
ILP solvers. Shorter clauses and PB constraints are much preferable
as they are easier to resolve against other constraints, and are more 
 useful to the learning strategies employed by exact SAT solvers. 
Another factor that gives instance-dependent SBPs the advantage is
the ease of symmetry detection, which was previously a bottleneck. 
Due to improved software \cite{Saucy}, the overhead of symmetry detection via
reduction to graph automorphism in SAT/0-1 ILP instances is almost
negligible. 

We also show that the three specialized 
0-1 ILP solvers, PBS II, Galena and Pueblo, all exhibit similar performance
 trends for different constructions. 
This indicates that performance
is not decided by solver-specific issues, but by the difficulty
of the instances and the SBPs added to them. CPLEX does not display the
same behavior as the other solvers, and is in fact slowed down
by the addition of instance-dependent SBPs and by several instance-independent
constructions. CPLEX is a commercial solver for 
generic ILP problems, and its algorithms and decision heuristics 
are likely to be very different than those used by academic solvers.
However, since details about CPLEX are not publicly available, 
it is not possible to accurately explain its behavior. We do note that
while CPLEX does not appear to benefit from symmetry breaking, 
its performance on the reduced instances with no SBPs of any kind
is superior to the 0-1 ILP solvers. However, once SBPs are
added the specialized solvers solve more instances than CPLEX in less time.

 In the context of generic search and combinatorial optimization problems defined in the 
 NP-spec language \cite{Cadoli99}, our empirical data
 suggest that new theoretical breakthroughs are required to
 make use of instance-independent symmetries during problem reductions
 to SAT or 0-1 ILP. At our current level of understanding,
 the simple strategy of processing instance-independent
 and instance-dependent symmetries together produces
 smallest runtimes for graph coloring benchmarks. Our current and
future work is focused on developing more effective SBPs for this problem,
and also investigating the utility of symmetry breaking
for other hard search  problems. Moreover, while our work uses instance-independent 
 predicates only for color symmetries, our results and analysis may have broader scope, for example, in applications 
such as radio frequency assignment (Section \ref{sec:bkg}) where symmetries are introduced during the reduction 
 to graph coloring and are likely to be preserved during future reductions. The issues involved in using instance-dependent
 vs. instance-independent SBPs are very relevant to such applications.

\section{Acknowledgments}
\label{sec:ack}
This work was funded in part by NSF ITR Grant \#0205288.
Also, we thank Donald Chai and Andreas Kuehlmann from UC Berkeley
for providing us with binaries of the Galena solver, and Hossein Sheini 
for providing us with the binaries for Pueblo.
\section*{Appendix A: Performance Analysis on {\tt Queens} Instances}
\begin{table*}[!t]
\scriptsize
\begin{center}
\begin{tabular}{|l||l|c|c|c|c|c|c|c|c|c|c|} \hline
& & \multicolumn{2}{c|}{PBS} & \multicolumn{2}{c|}{PBS II} & \multicolumn{2}{c|}{CPLEX} & \multicolumn{2}{c|}{Galena} & \multicolumn{2}{c|}{Pueblo} \\ \cline{3-12}
& & \multicolumn{2}{c|}{Inst.-dep.} & \multicolumn{2}{c|}{Inst.-dep.} &  \multicolumn{2}{c|}{Inst.-dep.} &  \multicolumn{2}{c|}{Inst.-dep.} &  \multicolumn{2}{c|}{Inst.-dep.} \\ 
Inst. & SBP & \multicolumn{2}{c|}{ SBPs used?} & \multicolumn{2}{c|}{ SBPs used?} &  \multicolumn{2}{c|}{ SBPs used?} & \multicolumn{2}{c|}{ SBPs used?} & \multicolumn{2}{c|}{ SBPs used?} \\ \cline{3-12}
Name & Type & No & Yes & No & Yes & No & Yes & No & Yes & No & Yes \\ \hline
 & no SBPs & T/O & {\bf 0.19} & 34.52 & 0.04 & 1.11 & 643.93 & 83.06 & 0.35 & 203.09 & {\bf 0.01} \\
& NU & {\bf 1.84} & T/O & 0.01 & 0.02 &  1.38 & 23.67 & {\bf 0.21} & {\bf 0.27} & 0.08 & 0.1 \\
queen5\_5 & CA & T/O & T/O & 0.31 & 0.24 & 39.2 & 2.76 & T/O & T/O & 0.14 & 0.52\\
& LI & 135 & 134.71 & 1.48 & 1.48 & 262.96 & 217.21 & 5.4 & 5.4 & 8.48 & 8.48  \\
& SC & 15.99 & 0.19 & 0.15 & 0.07 & {\bf 0.45} & 229.79 & 0.29 & 0.29 & 0.25 & 0.19 \\
& NU + SC & 8.63 & 12.34 & {\bf 0} & {\bf 0.01} &  0.83 & {\bf 0.88} & 0.3 & 1 &  {\bf 0.06} & 0.07 \\ \hline
 & no SBPs & T/O & 3.61 & T/O & 0.21 & T/O & T/O & T/O & {\bf 0.87} & T/O & 0.49 \\ 
& NU & 331.63 & 521.12 & 56.63 & 13.59 & T/O & T/O & 192.17 & 19.11 & 123.99 & 18.88 \\
queen6\_6 & CA & T/O & T/O & 50.6 & 780.57 & T/O & T/O & T/O & T/O & 196.94 & 80.53 \\
& LI &  T/O & T/O & T/O & T/O & T/O & T/O & T/O & T/O & T/O & T/O \\
& SC & T/O & {\bf 0.58} & T/O & {\bf 0.1} & 242.79 & T/O & T/O & 1.0 & T/O & {\bf 0.32} \\
& NU + SC & {\bf 2.89} & 1.72 & {\bf 1.4} & 0.63 & 95.91 & T/O & {\bf 11.19} & 1.05 & {\bf 4.85} & 2.64 \\ \hline
& no SBPs & T/O & 36.56 & T/O & 1.79 & 243.3 & T/O & T/O & T/O & T/O & {\bf 1.13} \\
& NU & {\bf 0.45} & {\bf 3.29} &  36.31 & 24.74 &  {\bf 119.16} & {\bf 459.44} & 56.6 & 147.52 & {\bf 9.59} & 15.49 \\
queen7\_7 & CA & T/O & T/O & T/O & T/O & 271.2 & T/O & T/O & T/O & 692.67 & 150.86 \\
& LI & T/O & T/O & 53.3 & 53.4 & T/O & T/O & 78.85 & 78.8 & 212.18 & 213.8 \\
& SC & T/O & 8.42 & 38.57 & {\bf 0.85} & 38.04 & T/O & T/O & {\bf 1.33} & 217.82 & 1.23 \\
& NU + SC & 5.65 & 38.07 & {\bf 4.37} & 5.73 &  119.7 & T/O &  {\bf 17.46} & 5.16 &  25.73 & 14.04 \\ \hline
& no SBPs & T/O & 1.31 & T/O & 0.52 & T/O & T/O & T/O & T/O & T/O & T/O \\
& NU & T/O & T/O & T/O & T/O & T/O & T/O & T/O & 138.61 & T/O & T/O \\
queen8\_12 & CA & T/O & T/O & T/O & T/O & T/O & T/O & T/O & T/O & T/O & T/O \\
& LI & T/O & T/O &  T/O & T/O & T/O & T/O & T/O & T/O & T/O & T/O \\
& SC & T/O & {\bf 1.05} & T/O & {\bf 0.47} & T/O & T/O & T/O & {\bf 1.9} & T/O & {\bf 0.98} \\
& NU + SC & T/O & T/O & {\bf 787.26} & 780.14 & T/O & T/O & {\bf 52.1} & 53.63 & T/O & T/O \\ \hline
\end{tabular}
\vspace{-2mm}
\parbox{5.8in} 
{
\caption{ \label{tab:queens} \small \bf
Detailed results for {\tt queens} instances. For each instance, 
we show results for the solvers PBS, PBS II, CPLEX, Galena and
Pueblo. All solvers are run on SunBlade 1000 workstations.
Instances are tested with no instance-independent SBPs,
with each of the four proposed constructions in Section
\ref{sec:sbconst} and with a combination of the NU and SC constructions.
All instance-independent SBPs are tested alone and with instance-dependent
SBPs added. The table shows the runtime for a given instance under
different construction. T/O indicates a timeout at 1000 seconds.
Best results for a given solver on each instance are shown in boldface.
}
}
\end{center}
\end{table*}

This section provides a more detailed discussion of our
results on individual benchmarks in the {\tt queens}
family of instances. The problem posed by {\tt queens}
instances is whether queens can be placed on
an $n \times m$ chessboard without conflicts. The instances
we use in our experiments are {\tt queens} $5 \times 5$,
$6 \times 6$, $7 \times 7$ and $8 \times 12$.
Table \ref{tab:queens} shows results for the {\tt queens}
family. Results are shown for every instance with no SBPs,
with each of the four constructions NU, CA, LI and SC, and
with the NU + SC combination. All constructions are tested
with and without instance-dependent SBPs as before. We report
results for the original version of PBS, from \cite{Aloul02b},
and for PBS II, CPLEX, Galena and Pueblo as in Section \ref{sec:results}.
Experiments are run on Sun Blade 1000 workstations as before.
In the table, we report solver runtime if an instance is solved,
and T/O for a timeout at 1000 seconds. The best results for
a solver on a particular instance are boldfaced.

While there is greater variation when considering performance
on a per-instance basis, the table largely reflects the same trends
reported in Section \ref{sec:results}. For example, when no instance-dependent
SBPs are used, PBS, PBS II, Galena and Pueblo all largely perform
best with the NU + SC construction. When instance-dependent
SBPs are added, the best performance is seen with the SC construction
in most cases. CPLEX does not display the same behavior as the other 
solvers, and its performance clearly deteriorates when instance-dependent
SBPs are added to any construction. A similar effect has been observed
in related work \cite{Aloul04}. Results for the original version of PBS \cite{Aloul02b},
which could not be included in Section \ref{sec:results}, have been added in
this section. It can be seen that PBS follows the same trends as 
PBS II, Galena and Pueblo, reinforcing our claim that this behavior is not
solver-dependent.

\vskip 0.2in
\bibliography{graphcol}
\bibliographystyle{theapa}

\end{document}